\documentclass[10pt,twocolumn,letterpaper]{article}

\usepackage{iccv}              % To produce the CAMERA-READY version

\usepackage{caption}
\usepackage{subcaption}
\usepackage{float}
\usepackage{placeins}
\usepackage{graphicx}
\usepackage{makecell}
\usepackage{stfloats}
\definecolor{iccvblue}{rgb}{0.21,0.49,0.74}
\usepackage[pagebackref,breaklinks,colorlinks,allcolors=iccvblue]{hyperref}

\def\paperID{13669} % *** Enter the Paper ID here
\def\confName{ICCV}
\def\confYear{2025}

\title{UniMLVG: Unified Framework for Multi-view Long Video Generation with Comprehensive Control Capabilities for Autonomous Driving}

\author{Rui Chen $^{1\dagger}$, Zehuan Wu $^{2\ast}$, Yichen Liu $^2$, Yuxin Guo $^2$, Jingcheng Ni $^2$, Haifeng Xia $^1$, Siyu Xia $^1$\\
$^1$ School of Automation, Southeast University, China 
$^2$ SenseTime Research\\
\{chenr, hfxia, xsy\}@seu.edu.cn; \{wuzehuan, liuyichen, guoyuxin, nijingcheng\}@sensetime.com \\
Code: \href{https://github.com/SenseTime-FVG/OpenDWM}{https://github.com/SenseTime-FVG/OpenDWM}
}

\begin{document}
\twocolumn[{%
\renewcommand\twocolumn[1][]{#1}%
\maketitle
\begin{center}    
    \centering    
    \includegraphics[width=\textwidth]{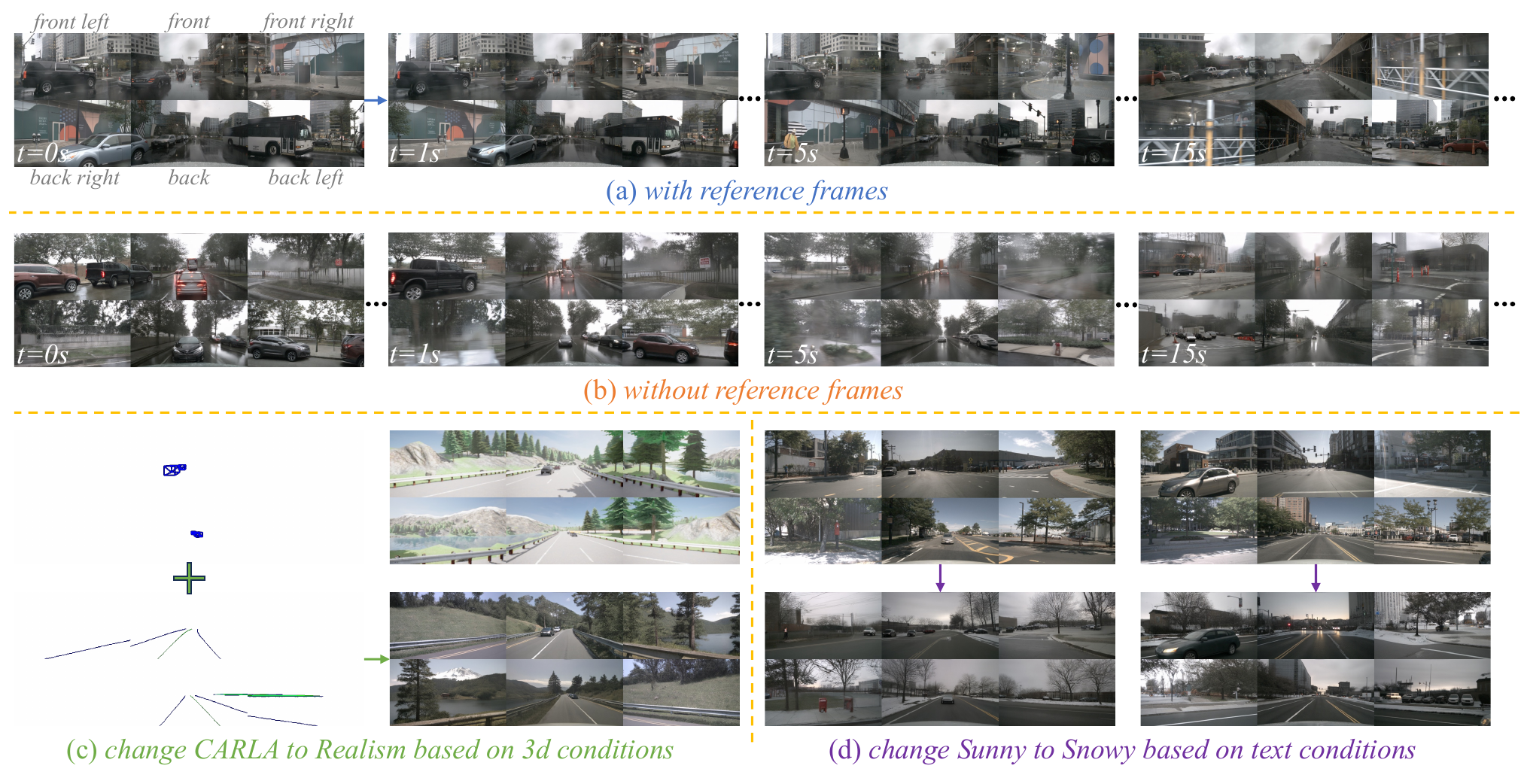}    
    \captionof{figure}{\textbf{Four tasks our model can perform}: (a) generating a $20$s multi-view video based on reference frames; (b) generating a $20$s multi-view video without any reference frames; (c) creating a realistic surround-view video from conditions obtained in a simulated environment; (d) altering weather conditions from sunny to snowy, driven by text-based prompts.}
    \label{fig:vis}
\end{center}%
}]
%\maketitle
 \footnotetext[1]{$^{\dagger}$ Work during the internship in SenseTime.}
 \footnotetext[2]{$^{\ast}$ Corresponding authors.}

%\begin{figure*}[hb]
%  \centering
%  \includegraphics[width=\textwidth]{fig/p1.pdf}
%  \caption{\textbf{Four tasks our model can perform}: (a) generating a $20$s multi-view video based on reference frames; (b) generating a $20$s multi-view video without any reference frames; (c) creating a realistic surround-view video from conditions obtained in a simulated environment; (d) altering weather conditions from sunny to snowy, driven by text-based prompts.}
%  \label{fig:vis}
%\end{figure*}

\begin{abstract}
The creation of diverse and realistic driving scenarios has become essential to enhance perception and planning capabilities of the autonomous driving system.
However, generating long-duration, surround-view consistent driving videos remains a significant challenge. To address this, we present UniMLVG, a unified framework designed to generate extended street multi-perspective videoscise control. By integrating single- and multi-view driving videos into the training data, our approach updates a DiT-based diffusion model equipped with cross-frame and cross-view modules across three stages with multi training objectives, substantially boosting the diversity and quality of generated visual content. 
Importantly, we propose an innovative explicit viewpoint modeling approach for multi-view video generation to effectively improve motion transition consistency. Capable of handling various input reference formats (e.g., text, images, or video), our UniMLVG generates high-quality multi-view videos according to the corresponding condition constraints such as 3D bounding boxes or frame-level text descriptions.ompared to the best models with similar capabilities, our framework achieves improvements of 48.2\% in FID and 35.2\% in FVD. 

% The rapid advancement of autonomous driving necessitates the creation of diverse and realistic driving scenarios to improve perception and planning capabilities. However, generating multi-view consistent videos with long duration remains challenging. To address this, we introduce UniMLVG, a unified framework for generating long-duration multi-perspective videos with precise and varied conditions. By integrating both single-view and multi-view driving videos into the training data, our approach updates cross-frame and cross-view blocks in two distinct stages and randomly drop these two kinds of blocks in training, whichc significantly enhances the stability of the model and the consistency of the generated outputs. UniMLVG has a comprehensive capability encompassing (unconditional generation, image-to-video generation, and video-to-video editing)
% nearly all generation and editing tasks required for autonomous driving video production, under conditions such as 3D bounding boxes and even frame-by-frame text descriptions. Our framework outperforms other state-of-the-art driving video generation methods, achieving the improvements of 30\% and 50\% in  (FID) and  (FVD), respectively.

\end{abstract}

\vspace{-7mm}
\section{Introduction}
\label{sec:intro}

Autonomous driving technology~\cite{uniad,jiang2023vad,yurtsever2020survey} is poised to transform human transportation and significantly enhance traffic safety. To achieve this, substantial data collection becomes necessary yet seriously increases both economic burden and labor costs. This situation motivates the exploration and adoption of simulation data. However, the disparity between simulated and real-world scenarios still obstructs practical application and perception algorithm.
For this issue, generative artificial intelligence provides a promising solution by synthesizing high-quality traffic data~\cite{yang2020surfelgan,gao2024vista,drive-wm}.
Meanwhile, the recent studies~\cite{drivedreamer, drivedreamer2} have demonstrated that these synthetic driving videos effectively support safe vehicle maneuvering, bridging critical gaps in real-world readiness.

Recent advancements in generating driving videos highlight several key requirements: long-term multi-view consistency, condition-based controllability, and diversity. However, existing generative algorithms fail to satisfy these requirements simultaneously. For instance, DriveGAN~\cite{drivegan} and DriveDreamer~\cite{drivedreamer} introduce action-based autoregressive techniques to generate next-frame driving images. Nevertheless, they both overlook the importance of multi-view perspectives in autonomous driving, limiting their ability to generate comprehensive multi-view data. MagicDrive~\cite{magicdrive} addresses this by incorporating 3D control information and cross-view modules to create multi-view images. Although they leverage the approach~\cite{wu2023tune} to produce videos, the outputs are short in duration and lack temporal consistency. More recent efforts~\cite{drive-wm, mei2024dreamforge, dive} have shifted focus toward generating controllable, multi-view videos over longer sequences. However, due to the constraints of single-objective training and small-scale datasets, these methods still struggle with achieving both temporal consistency and diversity. Notably, all of these works rely on scene-level descriptions to generate corresponding videos, but they neglect the inclusion of fine-grained textual conditions, which limits their overall quality and controllability.

To address the limitations outlined above, this paper introduces a novel generative framework, UniMLVG, designed to generate highly consistent and controllable multi-view long videos. Building on a DiT-based image generation model~\cite{sd3}, UniMLVG integrates temporal and cross-view modules to capture dynamic sequences and multi-view information. To mitigate error accumulation caused by long-term autoregressive generation, we propose a multi-task training objective. Additionally, to ensure consistent motion across views, we introduce explicit perspective modeling. Unlike previous methods that directly encode camera intrinsic and extrinsic parameters, our approach encodes camera rays within a unified spatial framework, injecting spatially informed physical knowledge. For enhancing the diversity of generated instances, we leverage over 1,000 hours of driving scene data for model training.
Regarding multi-condition control, we integrate 3D conditions—such as 3D bounding boxes and high-definition maps (HDmaps)—alongside image-wise textual descriptions.
In conclusion, UniMLVG effectively supports text-to-video (T2V), image-to-video (I2V), and video-to-video (V2V) generation, demonstrating exceptional performance in both text-based and 3D condition-based editing, as illustrated in Fig.~\ref{fig:vis}. The primary contributions:
\begin{itemize}
    \item To simulate the practical traffic scenarios, a novel multi-task, multi-condition, multi-stage training strategy is developed into UniMLVG and improves training stability.
    \item To enhance the motion coherence and the consistency across multiple views, we inject spatially physical knowledge through explicit perspective rays modeling.
    \item Extensive empirical analyses reveal that leveraging multiple datasets and image-level descriptions significantly enhances the diversity and text-based controllability of generated videos across various weather and times.
    \item According to experimental results, our UniMLVG surpasses the existing street video generation techniques in temporal consistency and frame quality, especially in its capability to perform  diverse generation tasks.
\end{itemize} 
\section{Related Works}
\label{sec:related}
\subsection{Video Generation and Editting}

Video generation and editing hold vital importance in various fields, particularly in autonomous driving. These technologies enable the creation and manipulation of realistic video content, which is crucial for improving the perception and planning of autonomous systems in diverse scenarios. 
Traditional methods for video generation widely 
adopt techniques including Autoencoder~\cite{videovae, denton2018stochastic, hsieh2018learning}, Generative Adversarial Network (GAN)~\cite{tganv2, vid2vid, styleganv, dvdgan, storygan} and Autoregressive Model~\cite{yan2021videogpt, harp, nuwa, godiva}. However, these approaches often fail to create realistic and diverse video and are unable to precisely control the content through text or layout. 

In recent years, the advancement of diffusion models has revolutionized the field of image and video generation. Integrating the aforementioned techniques into the diffusion models, diffusion-based approaches~\cite{videodiffusionmodel, latte, yuan2024instructvideo, zhao2024real, tang2024any, guo2023animatediff, fatezero, motionctrl, champ} has become the mainstream of video generation and editing, due to their ability to produce high-quality videos with comprehensive control capabilities. 
% Inital works like~\cite{videodiffusionmodel} adopt 3D UNet to create videos with fixed-frame and generate the following frames autoregressively conditioned on the previous generation. Some improvement such as~\cite{latte} conduct the diffusion process in the latent space and use different attention blocks to capture spatiotemporal distribution of videos. 
Furthermore, the promising results of large-scale generative models like~\cite{sora, Open-Sora} have inspired researchers to use these models as real-world simulators, significantly influencing the field of driving simulation~\cite{gaia1, drivedreamer, opendv}. For instance, DriveDreamer~\cite{drivedreamer} takes the reference frame, the road structural information and text description as input and employs three types of attention blocks within the diffusion mode to predict the future frames. GenAD~\cite{opendv} leverages a large scale of YouTube videos as data to pretrain data and divides the training process into two stages, allowing the model to progressively learn image and video denoising. Despite their success, these methods generate single-view videos, which are less useful compared to multi-view videos, as autonomous vehicles need to perceive the surroundings rather than just the front view information only.

\subsection{Multi-view Video Generation}
Compared to standard video generation, multi-view video generation remains relatively underexplored, due to the challenges in ensuring the consistency across perspectives and time series initial trials simply adopts the video diffusion architecture. For instance, MagicDrive~\cite{magicdrive} encodes high-level controls such as text and bounding boxes independently and flatten those conditions into an embedding sequence. In contrast, DreamForge~\cite{mei2024dreamforge} employs a ControlNet~\cite{controlnet} to fuse multi-modality condition.
Drive-WM~\cite{drive-wm} generates videos from a few views using a diffusion model in the first stage and then utilizes multi-view factorization to predict additional views conditioned on the pre-generated ones.
DiVE~\cite{dive} leverages the DiT structure within its diffusion model and applies a view-inflated attention mechanism to compute attention across features from all viewpoints.
Extending the design of~\cite{drivedreamer}, DriveDreamer-2~\cite{drivedreamer2} concatenates all images per frame into a large single image, and only cross-frame modules are adopted in the forward pass.
Despite these approaches, they usually rely on a reference frame to generate subsequent content and the quality of the videos significantly degrades as the sequence length increases. Additionally, they struggle to address inconsistencies across views and timestamps.
Based on the aforementioned issues,  we introduce our unified model, which can provide precise control, generate high-quality videos and guarantee the spatial and temporal consistency.  
\section{Method}
\label{sec:method}

\begin{figure*}[t]
\centering
    \vspace{-4mm}
    \includegraphics[width=0.99\linewidth]{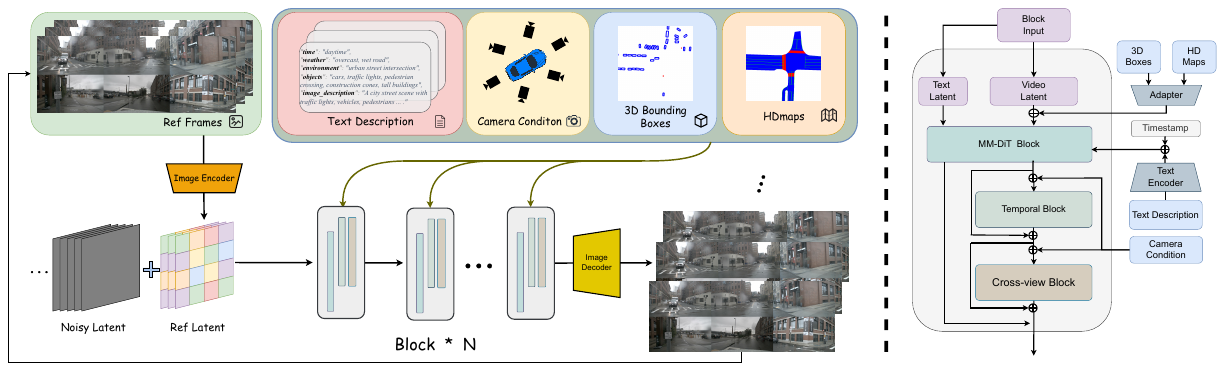}
    \vspace{-4mm}
    \caption{\textbf{Overall framework of the model.} \textit{left}: The encoded reference frames are concatenated with the noisy latent as video latent and fed into $N$ UniMLVG blocks. The diverse conditions including image-level descriptions, camera pose, and 3D conditions are injected into each UniMLGV block and interact with the video latent to guide the generated contents. Finally, the model outputs the subsequent frames, which can then be used as the reference frame for the next autoregressive generation. Note that our model can produce driving video based on those conditions only, where the reference frames are not required. \textit{right}: Details of the UniMLVG block. A UniMLVG block comprises three distinct sub-blocks to perform attention across different dimensions, while the different conditions are integrated into the video latent in different positions during the forward passing.
    % The model's inputs include the noise latent, latent of reference frames, image-level descriptions, camera pose conditions, and 3D conditions. After passing through $N$ blocks, the model generates the predicted frame, which is then used as the reference frame for the next autoregressive step. The detailed structure of the block is shown in the figure on the right.
    }
    \vspace{-4mm}
    \label{fig:model}
\end{figure*}

Fig.~\ref{fig:model} illustrates the overall architecture of our proposed UniMLVG. Our method enhances the DiT-based image generation model~\cite{sd3} with two additional modules: the temporal module $\mathcal{T}$ and the cross-view module $~\mathcal{C}$. 
% Using a multi-task, multi-condition, and multi-stage training strategy across diverse datasets, 
UniMLVG is capable of generating extended street-view videos and performing text-based and 3D-conditioned editing by using a multi-task, multi-condition and multi-stage training strategy across diverse datasets. Importantly, we introduce explicit perspective modeling for the first time to incorporate physical spatial information, enabling smoother and more consistent multi-view autonomous driving scene videos.

\subsection{Unified Framework}
We choose the DiT-based model~\cite{sd3} as the backbone for image generation over alternatives like~\cite{sd, svd, flux, Open-Sora} and extend the MM-DiT block, shown in the rigth of Figure~\ref{fig:model} to our UniMLVG block to achieve an optimal balance of generation quality, scalability, model size, and flexibility in text-based control. For our diffusion model's optimization, we employ the conditional rectified flows~\cite{lipman2022flow, liu2022flow, sd3} loss as the primary objective:
\begin{equation}
\mathcal{L} = \mathbb{E}_{\epsilon \sim \mathcal{N}(0, I)} \left[ \left\| v_{\theta}(z_t, t, c) - (z_0 - \epsilon) \right\|^2 \right],
\end{equation}
where $v$ is the model, $z$ is the video latent, $\theta$ is the learnable parameters, $t$ is the timestep and $c$ is the condition.

The temporal and cross-view modules are positioned directly after the Multimodal Diffusion Transformer (MM-DiT) block to maintain consistency across time and viewpoints. Specifically, we apply the same self-attention module while adjusting input dimensions to suit the specific task. Given the noisy latent $z_{t}\in \mathbb{R}^{T \times V \times H \times W \times C}$ 
, where $T$ is the frame length and $V$ represents the number of viewpoints, we adjust the attention sequence lengths differently for the temporal and cross-view modules. In the temporal module, we flatten the dimensions $VHW$, resulting in a shape of $T \times (VHW) \times C$. In contrast, in the cross-view module, we combine $THW$, producing $V \times (THW) \times C$. Additionally, we use an absolute positional encoding method~\cite{vaswani2017attention} that employs sine and cosine functions at varying frequencies to separately encode the positional information of both frames and viewpoints.

\subsection{Multi-task}
We observe that when using only the video prediction task, the quality of the generated video deteriorates significantly after a few autoregressive iterations. We believe this is due to the model's excessive reliance on reference frames, which leads to accumulated autoregressive errors. To address this issue, we propose a multi-task training strategy to improve long-term video quality and coherence. Specifically, we design four training tasks: video prediction (VP), image prediction (IP), video generation (VG), and image generation (IG). In VP, we utilize the embeddings of the first $k$ frames from $n$ viewpoints as reference frames, setting their timestep to $0$. In IP, we mask 50\% of the reference frames, requiring the model to use the remaining discrete reference information to generate images at the masked positions.
In VG, the model generates the next $l$ frames of a multi-view video based solely on the given conditions, without any reference frames. In IG, we drop the temporal module to prevent the model from overly relying on temporal continuity, thus preserving its ability to maintain cross-view consistency. During training, the first three tasks are executed by sampling a mask $M \in \mathbb{R}^{T \times V}$ based on a predefined ratio, with the loss calculation excluding predictions of reference frames, as described below:
\begin{equation}
\mathcal{L} = \mathbb{E}_{\epsilon \sim \mathcal{N}(0, I)} \left[ \left\| (1-M) \odot ( v_{\theta}(z_t, t, c) - (z_0 - \epsilon)) \right\|^2 \right].
\end{equation}

Notably, unlike DriveDreamer-2~\cite{drivedreamer2}, which considers only the VP and VG tasks, UniMLVG additionally incorporates the IP and IG tasks. As shown in Tab. \ref{tab:mt} and Fig. \ref{fig:abl_vp}, this contributes significantly to improving the quality of generated frames and mitigating autoregressive errors.

\subsection{Multi-condition}
To help the model grasp the physical dynamics of autonomous driving scenes, we introduce both local conditions (such as 3Dboxes and HDmaps) and global conditions (including view-specific text descriptions). Notably, we are the first to explicitly model camera parameters to incorporate physical spatial information for this task.

\noindent\textbf{Local Conditions.} We unify local conditions as image-based conditions. Similar to the approach used in T2I-Adapter~\cite{t2i-adapter}, we introduce a lightweight image adapter for efficient processing. By projecting 3Dboxes or HDmaps and mapping the instance identity to the color space, we generate sparse images $I_l \in \mathbb{R}^{T\times V\times H'\times W'\times 3}$, matching the original image size. These conditional images are concatenated and input into the adapter to obtain multi-level features $C_l \in \mathbb{R}^{k\times T\times V\times H\times W\times C}$, which are then added to the latent representation at the corresponding layer. 

\noindent\textbf{Global Conditions.} Building on the approach in~\cite{sd3}, we utilize three text encoders~\cite{clip, openclip, t5} to extract multi-level textual features, which are managed through joint attention mechanisms and AdaLN~\cite{perez2018film} with latent variables. 
Notably, we use image-level descriptions instead of scene-level ones, enabling more customized content generation. In our method, the primary global condition is text, but the framework is also well-suited to incorporate additional information such as video frame rate and actions. 

\begin{figure}[t]
  \centering
   \includegraphics[width=1\linewidth]{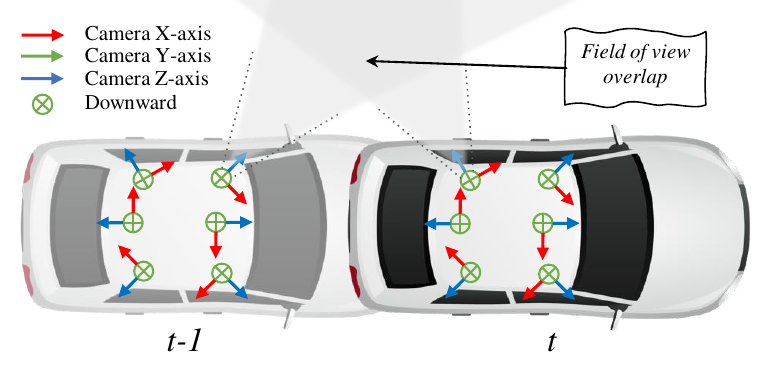}
   \vspace{-6mm}
   \caption{\textbf{Field of view overlap between cameras over time.}} 
   % It can reduce the uncertainty in generating new scene frames.}
   \label{fig:campos}
   \vspace{-5mm}
\end{figure}

\noindent\textbf{Explicit Perspective Modeling.} Apparently, there exist overlaps in the image contents between the adjacent views at the same moment and between the same view at different time. Moreover, as shown in Figure~\ref{fig:campos}, there is also a significant field of view overlap between the current viewpoint of the moving vehicle and one of previous viewpoint. These overlaps help reduce uncertainty in generating scene images at new time steps. Meanwhile, modeling camera poses explicitly has proven effective in recent works~\cite{cat3d, scene} that generate multi-view images of static scenes. We believe this is also applicable for generating consistent dynamic driving scenes. Specifically, we set the optical center of the forward-facing camera in the first frame as the origin to establish a unified coordinate system for all viewpoints across different time steps. We encode the camera ray representation \cite{sajjadi2022scene} to produce ray features, which are added as spatial information to the video latent. The formula for ray coordinate encoding is as follows:
\begin{align*}
\mathcal{P}(c) = &[\sin(2^0 \pi c), \cos(2^0 \pi c), \sin(2^1 \pi c), \\
             &\cos(2^1 \pi c), \dots, \sin(2^{j-1} \pi c), \cos(2^{j-1} \pi c)] \\
Enc(\mathbf{v}) = &MLP(Concat(\mathcal{P}(\mathbf{v}_0),\mathcal{P}(\mathbf{v}_1), \dots, \mathcal{P}(\mathbf{v}_k))) ,
\end{align*}
where $\mathcal{P}$ denotes the encoding function applied to each value of $\mathbf{v}$ (the center and direction of ray), $j$ represents the encoding dimension (set to $8$ in this paper).

\begin{table*}
\centering
\begin{tabular}{l|cccccccc}
\toprule
\textbf{Method} & \textbf{Multi-view} & \textbf{Video} & \textbf{Duration} & \textbf{FID}$\downarrow$ & \textbf{FVD}$\downarrow$ & $\textbf{mAP}_{obj}$$\uparrow$ & $\textbf{mIoU}_{road}$$\uparrow$ & $\textbf{mIoU}_{vehicle}$$\uparrow$\\ 
\midrule
\midrule
Oracle & - & - & - & - & - & 35.56 & 73.67 & 31.86 \\ 
\midrule
DriveGAN~\cite{drivegan} & $\times$ & $\surd$  & 3s & 73.4 & 502.3 & - & - & - \\ 
DriveDreamer~\cite{drivedreamer} & $\times$ & $\surd$ & 4s& 52.6 & 452.0 & -& - & -\\
MagicDrive~\cite{magicdrive} & $\surd$ & $\surd$ & 5s & 19.1 & 218.1 & 12.30 & 61.05 & \underline{27.01}\\ 
Drive-WM~\cite{drive-wm} & $\surd$ & $\surd$ & 20s & 15.2 & 122.7 & 20.66 & \underline{65.07} & - \\ 
DriveDreamer-2~\cite{drivedreamer2} & $\surd$ & $\surd$ & 7s & \underline{11.2} & \underline{55.7} & - & - & -\\ 
DreamForge~\cite{mei2024dreamforge} & $\surd$ & $\surd$ & 20s & 16.0 & 224.8 & 13.80 & - & - \\ 
DiVE~\cite{dive} &$\surd$ & $\surd$ & 20s & - & 94.6 & \textbf{24.55} & - & -\\ 
\midrule
\textbf{Ours} & $\surd$ & $\surd$ & 20s & \textbf{5.8} & \textbf{36.1} & \underline{22.50} & \textbf{70.81} & \textbf{29.12}  \\ 
\bottomrule
\end{tabular}
\caption{\textbf{Comparison of the generation quality and condition-following metrics on nuScenes validation set.} $\uparrow$/$\downarrow$ indicates that a higher/lower value is better. The best results are in \textbf{bold}, while the second best results are in \underline{underlined} (when other methods are available).}
\vspace{-0.2in}
\label{tab:res}
\end{table*}

\subsection{Multi-stage}
UniMLVG utilizes a wide range of currently available datasets, including both single-view and multi-view street scene videos, to impart driving scene priors to the model. 
Considering that single-view and multi-view data differ in the number of viewpoints and annotation information, we implement a multi-stage training strategy to ensure stable and efficient model convergence. It is worth noting that our multi-stage training does not follow the typical approach of first training on images and then fine-tuning on videos as in previous works. Instead, all stages of our training are conducted using video data.

\noindent\textbf{Stage I.} Empowering the model with the capability to anticipate future driving scenarios from a forward-facing perspective. We train $\mathcal{T}$ on a substantial collection of publicly available forward-facing driving videos~\cite{opendv}. This dataset, notable for its large-scale, high-resolution and multi-scenario, helps the model generate temporal coherent frames. During this phase, we freeze the initial weights of SD3~\cite{sd3} and bypass the cross-view modules.

\noindent\textbf{Stage II.} Infusing the model with the ability to generate from multiple viewpoints and effectively follow conditional inputs. We train $\mathcal{C}$ and $\mathcal{T}$ using several multi-view datasets~\cite{nuscenes, waymo, argoverse}. These datasets provide multi-view videos, camera calibration parameters, and 3D annotations. During this phase, only the temporal and cross-view modules are trained, while the backbone remains frozen.

\noindent\textbf{Stage III.} To further improve generation quality, we perform full fine-tuning in this phase. With the model's generation capabilities well-developed from the first two stages, this phase typically requires 1 or 2 training epochs.

\section{Experiments}
\label{sec:rexperimental}

\subsection{Experiment Details}
\noindent\textbf{Datasets.} We use the single-view dataset OpenDV-Youtube~\cite{opendv} and the multi-view datasets nuScenes~\cite{nuscenes}, Waymo~\cite{waymo}, and Argoverse2~\cite{argoverse}. OpenDV-Youtube is exclusively used for Stage I, while nuScenes, Waymo, and Argoverse2 are divided into training and validation sets following their original splits. The training duration totals $1, 498$ hours, comprising $1,486$ hours from OpenDV-Youtube, $4.6$ hours from nuScenes, $4.4$ hours from Waymo, and $3.1$ hours from Argoverse2. 
We leverage available dataset annotations, including 3Dboxes, HDmaps and camera parameters. Meanwhile, nuScenes with $12$ Hz interpolated annotations~\cite{nusc_an} is used. 
Additionally, text descriptions for all frames and views are generated at $2$ Hz using~\cite{drivemlm}.

\noindent\textbf{Evaluation Metrics.} To assess the effectiveness of our method in terms of realism, continuity, and precise control, we selected four key metrics to compare against existing multi-view image and video generation methods. For realism, we use the widely recognized Fréchet Inception Distance (FID)~\cite{fid}. To estimate temporal coherence in our videos, we measure consistency using Fréchet Video Distance (FVD)~\cite{fvd}.  For a fair comparison, we conduct evaluation following~\cite{magicdrive}, using $150$ scenes from the nuScenes validation set, with $6$ viewpoints per scene and $16$ frames per viewpoint, totaling $900$ videos for FVD and FID computation. 
% Additionally, we sample $10,000$ images from these $900\times 16$ frames to calculate FID. 
For controllability, we evaluate two perception tasks: 3D object detection~\cite{bevfusion} and BEV segmentation~\cite{zhou2022cross}, following the approach MagicDrive~\cite{magicdrive}.

\noindent\textbf{Implement Details.} We use $3$ frames as reference for autoregressive prediction. A fixed learning rate of $8 \times 10^{-5}$ is applied across all stages and optimized with AdamW~\cite{adamw}. The conditions dropping rate is set uniformly at $20\%$, and classifier-free guidance scale is $3$. The inference steps set $50$. All experiments are conducted on A$800$ GPUs.

\subsection{Experiment Results}

\textbf{Quantitative Results.} We report quantitative experimental metrics on the nuScenes validation set, as shown in Tab.~\ref{tab:res}. Overall, our model achieves quite promising results, with a significant improvement of 48.2\% in FID and 35.2\% in FVD compared to the second-best method. As demonstrated in the ablation study~\ref{sec:abl}, the multi-task, multi-stage training strategies and explicit perspective modeling make significant contributions. In terms of temporal consistency, we achieve a 61.8\% improvement over DiVE~\cite{dive}, which also uses the DiT~\cite{videodiffusionmodel} architecture. Compare to DriveDreamer-2~\cite{drivedreamer2}, UniMLVG achieve a 35.2\% improvement in FVD. Moreover, DriveDreamer-2 can only produce 7s video while UniMLVG can generate longer video up to $20$s, almost three times as longer as DriveDreamer-2. Additionally, unlike DriveDreamer-2~\cite{drivedreamer2}, UniMLVG does not stitch multiple views into a single image for generation, which reduces memory usage and training time. On contrast, our cross-view module can easily handle the different numbers of viewpoints presented in different datasets using view mask. In terms of condition adherence, UniMLVG achieve the SOTA results, improving $\text{mIoU}_{road}$ and $\text{mIoU}_{vehicle}$ by 8.8\% and 7.8\%, respectively, over the second-best method. For mAP, we outperform all other methods except DiVE~\cite{dive}. 
In contrast to DiVE, UniMLVG does not assign distinct classifier-free guidance scales for each condition, significantly reducing inference time. Furthermore, rather than using the parameter-heavy ControlNet~\cite{controlnet}, UniMLVG employs a lightweight adapter, enabling more flexible and efficient condition integration.

\begin{figure*}[t]
\centering
    \includegraphics[width=1\linewidth]{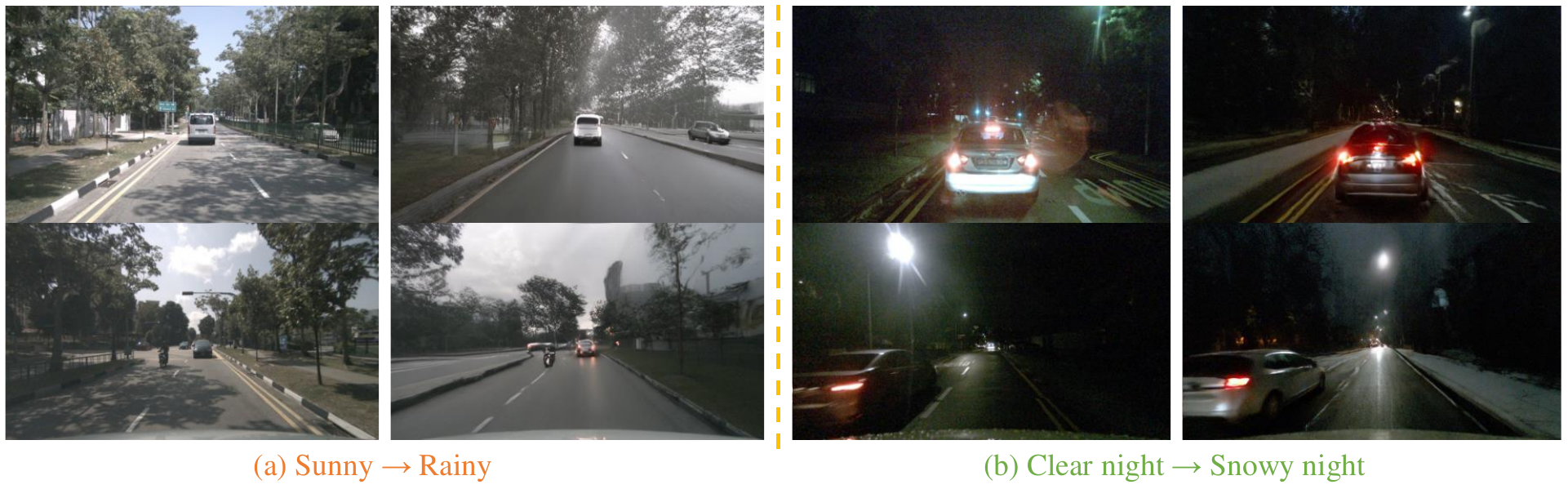}
    \vspace{-6mm}
    \caption{
    \textbf{Text-based weather editing at different times of day}: (a) shows text-based control changing sunny to rainy. (b) demonstrates text editing to generate a snowy night scenario. In each subfigure, the left side shows the ground truth, while the right side presents the generated results, with the top and bottom representing the front and rear viewpoints.}
    \label{fig:text_cond}
    
\end{figure*}

% \begin{figure*}[t]
%   \centering
%    \includegraphics[width=1\linewidth]{fig/model.png}
%    \caption{Framework.}
%    \label{fig:framework}
% \end{figure*}
\begin{table}[h]
    \centering
    \begin{tabular}{cccc|cc}
        \toprule
        \textbf{3Dbox} & \textbf{HDmap} & \makecell{\textbf{Cam.}\\\textbf{id.}} & \makecell{\textbf{Cam.}\\\textbf{ray.}} & \textbf{FVD$\downarrow$} &       \textbf{FID$\downarrow$} \\
        \midrule
          & & & &  242.46 & 34.55 \\
          $\surd$  & & & & 49.30 & 8.80 \\
          & $\surd$ & & & 49.18 & 8.86 \\
          $\surd$  & $\surd$ & &  & 50.13 & 8.91  \\
          $\surd$  & $\surd$ & $\surd$ & & 50.76 & 8.74  \\
          $\surd$  & $\surd$ &  & $\surd$ & \textbf{44.43} & \textbf{6.78} \\
        \bottomrule
    \end{tabular}
    \caption{\textbf{Ablation studies on 3D conditions and camera pose modeling.} Cam. id. and Cam. ray. represent implicit and explicit modeling, respectively.}
    \vspace{-0.2in}
\label{tab:cond}
\end{table}
\noindent\textbf{Controllability.} 
Our model supports 3D condition control as well as text-based control capabilities. In Figure~\ref{fig:vis}(c), we generate realistic scenes using 3D conditions obtained from the simulation engine CALAR~\cite{dosovitskiy2017carla}. Notably, different from style transfer methods, our generation is not strictly constrained to transforming the content of simulated scenes. Instead, our model takes the general 3D information such as 3D bounding boxes and HDMaps as inputs and leverages real-world knowledge to generate plausible scenes that align with the distribution of the training data. Specifically, elements such as roads, mountains, and trees are generated with appearances that more accurately reflect how they would look in the real world. For textual control, the model illustrates strong cross-distribution transfer and impressive editing capabilities under extreme transformations. As shown in Figures~\ref{fig:vis}(d) and~\ref{fig:text_cond}, even though the nuScenes dataset does not contain any snowy, our model can still edit the weather of nuScene multi-view videos to snowy in both daytime and nightime. Additionally, the generated videos accurately present real-world details such as snow accumulating on the roadside, bare trees during winter, and reflective road surfaces due to rain.

\begin{figure*}[t]
\centering
    \includegraphics[width=1\linewidth]{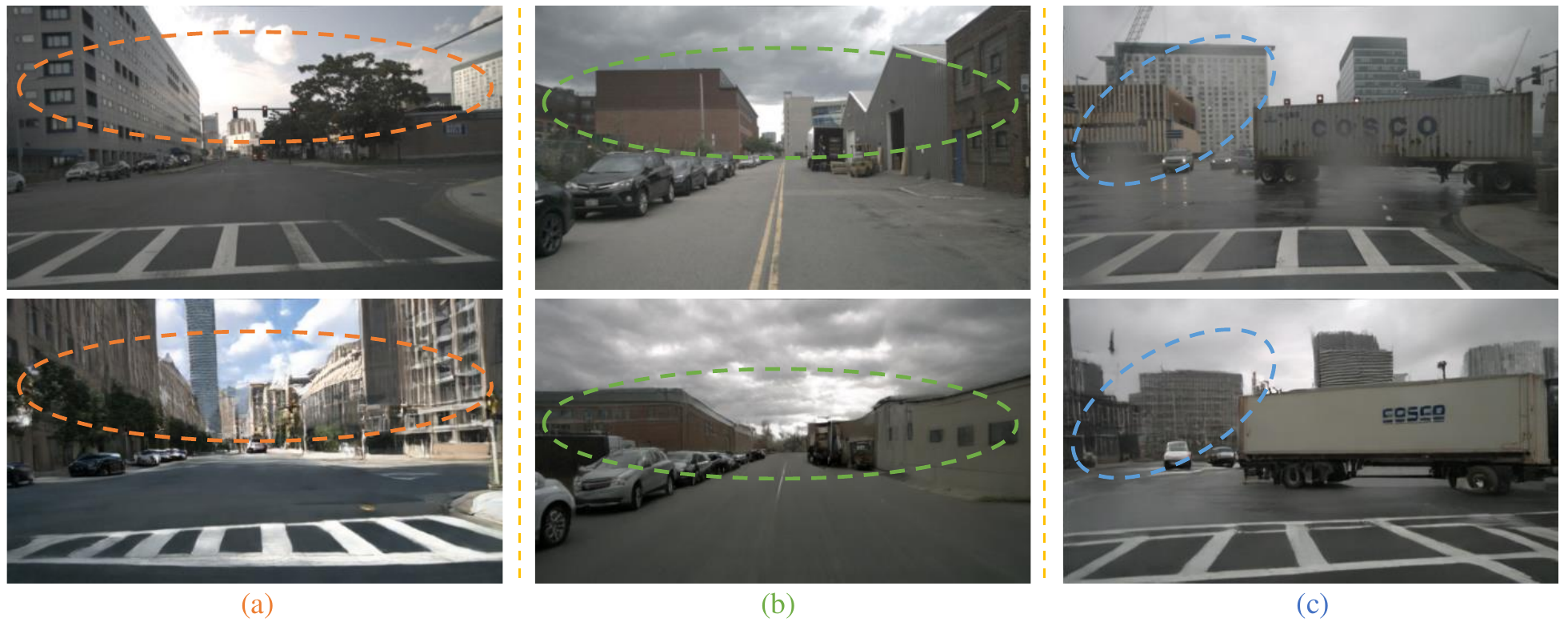}
    \vspace{-0.3in}
    \caption{\textbf{Examples of scene generation diversity under various weather conditions.} (a) Under sunny conditions, the appearance and number of houses, cloud positions, and sunlight direction differ from the ground truth (GT). (b) Under cloudy conditions, the appearance of houses and the colors of nearby vehicles differ from GT. (c) Under rainy conditions, both the appearance of houses and vehicles deviate from GT. The top row displays the ground truth.}
    \label{fig:show}
    \vspace{-0.25in}
\end{figure*}
\begin{figure}[t]
\centering
    \includegraphics[width=0.99\linewidth]{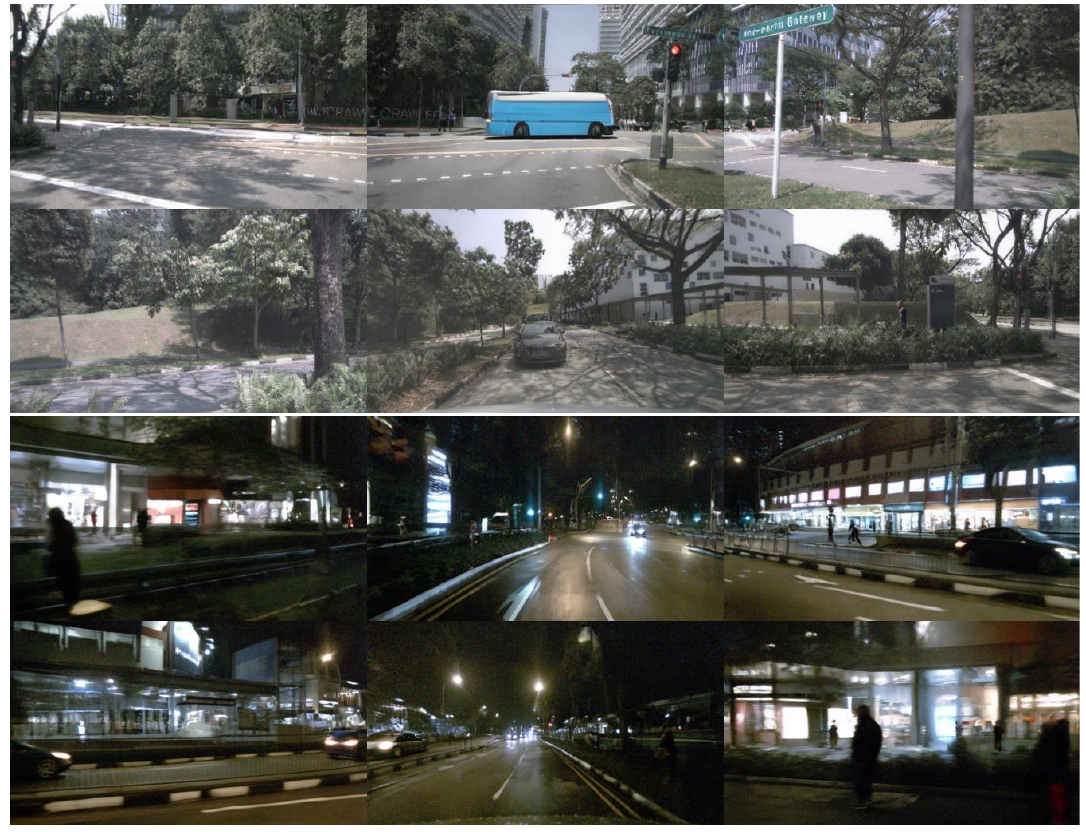}
    \vspace{-0.15in}
    \caption{\textbf{Examples of scene generation consistency.} The lane markings and bushes at the junction of viewpoints remain consistent both during the day and at night.}
    \label{fig:cons}
    \vspace{-0.125in}
\end{figure}
\noindent\textbf{Diversity \& Consistency.} 
One primary consideration of driving video generation mdoels is the diversity of the results, as this task is fundamentally intended to address the scarcity of annotated data. In Figure~\ref{fig:show}, we provide examples to showcase the diversity of our model's output. 
% {\blue lyc: You can describe some details in the caption of that figure and save some space here: can observe that, compared to the ground truth, our generated results show differences in buildings, sky, and vehicles. For example, the shape, color, and number of buildings (Fig.~\ref{fig:show} (a)(c)), as well as the shape and color of vehicles (Fig.~\ref{fig:show} (b)), may change. }
The surroundings such as weather, buildings and cars varies from cases to cases, while the generated results strictly follow the real conditions in terms of road layout and vehicle positions. 
Providing multi-view consistent videos helps enhance the perception capabilities of autonomous driving algorithms. In Figure~\ref{fig:cons}, we present examples of multi-view consistency under both daytime and nighttime conditions. We can observe that lane markings and vegetation remain seamlessly continuous at the boundaries between viewpoints.
Please refer to the supplementary materials for more diverse and consistent examples.

\subsection{Ablation Studies}
\label{sec:abl}
We conduct a series of ablation studies to evaluate the distinct contributions of multi-condition, multi-task, multi-dataset and multi-stage training. All the following evaluation metrics are reported on the nuScenes validation set. Unless otherwise specified, the ablation experiments are conducted only on Stage 2. Additionally, the weights obtained from the Stage 1 are used as the initial weights for all experiments to enhance training efficiency.

\begin{table}[t]
    \centering
    \begin{tabular}{ccc|cc}
        \toprule
        \textbf{N} & \textbf{O} & \textbf{W\&A} & \textbf{FVD$\downarrow$} &       \textbf{FID$\downarrow$} \\
        \midrule
        $\surd$ & &  & 58.26 & 8.83 \\
        $\surd$ & $\surd$ & & 52.52 & 8.71 \\
        $\surd$ & $\surd$ & $\surd$ &\textbf{44.43} & \textbf{6.78} \\
        
        \bottomrule
    \end{tabular}
    \caption{\textbf{Ablations of Multi-dataset.} N, O, W, A represent nuScenes, OpenDV-Youtube, Waymo and Argoverse2, respectively.}
\label{tab:opendv}
\end{table}
\begin{table}[t]
    % \large
    \centering
    \begin{tabular}{l|cc}
        \toprule
        \textbf{Stage} & \textbf{FVD$\downarrow$} &       \textbf{FID$\downarrow$} \\
        \midrule
        Stage I    & 149.70 & 30.50 \\
        Stage II    & 44.43  & 6.78  \\
        Stage III    & \textbf{36.11} & \textbf{5.82} \\
        \bottomrule
    \end{tabular}
    \caption{\textbf{Ablations Studies of Multi-stage.} Our multi-stage training strategy can significantly improve the generation quality and consistency.}
    \vspace{-0.1in}
\label{tab:mts}
\end{table}
\noindent\textbf{Multi-dataset.} Table~\ref{tab:opendv} presents the training results with and without integrating datasets other than the nuScenes~\cite{nuscenes} dataset. The results clearly demonstrate that leveraging a large amount of single-view data significantly enhances the continuity and quality of the videos. This indicates that substantial unlabeled single-view street scene videos can effectively enhance the model's ability to imagine and predict future scenes. Additionally, by integrating two multi-view video datasets for joint training, we achieved a 15.4\% improvement in FVD. This demonstrates the model's scalability and highlights the critical role of diverse data in enhancing its understanding of driving scenes.

\noindent\textbf{Multi-condition.} Based on the ablation results from multiple datasets, we use the weights trained on the OpenDV-Youtube as initialization for the subsequent experiments. We compare the results for different combinations of the four conditions, as shown in Table~\ref{tab:cond}. We find that incorporating 3D conditions lead to substantial improvement in FVD and FID, as the 3D information helps the model interpret the driving scene. In addition, we also conduct the comparison between implicit (Cam.id.) and explicit (Cam.ray.) camera pose modeling. Cam.id. used in ~\cite{magicdrive} encodes the camera's intrinsic and extrinsic parameters into a vector, which is handled as global information like textual inputs. The results show that the explicit modeling approach not only improves video continuity by 12.5\% but also enhances image quality by 22.4\%. This suggests that explicitly incorporating positional relationships between viewpoints into the model enhances its understanding of driving scenes. Additionally, the implicit modeling approach does not improve the model’s performance.

\begin{table*}[t]
    \centering
    \begin{subtable}[t]{0.33\linewidth}
    \begin{tabular}{cc|cc}
        \toprule
        \textbf{VP}(\%) & \textbf{IP}(\%) & \textbf{FVD$\downarrow$} &       \textbf{FID$\downarrow$} \\
        \midrule
        100 & 0    & 52.52 & 8.71 \\
        95 & 5    & 45.46 & \textbf{8.03} \\
        90 & 10    & 48.71 & 8.91 \\
        80 & 20    & 50.74 & 8.93 \\
        70 & 30    & \textbf{40.05} & 8.45 \\
        60 & 40    & 46.26 & 8.52 \\
        \bottomrule
        \end{tabular}
        \caption{Different ratios of VP and IP.}
    \end{subtable}
    \hfill
    \begin{subtable}[t]{0.33\linewidth}
    \begin{tabular}{cc|cc}
        \toprule
        \textbf{VP}(\%) & \textbf{VG}(\%) & \textbf{FVD$\downarrow$} &       \textbf{FID$\downarrow$} \\
        \midrule
        100 & 0    & 52.52 & 8.71 \\
        95 & 5     & 47.30 & \textbf{8.54} \\
        90 & 10    & \textbf{40.88} & 8.66 \\
        80 & 20    & 53.50 & 9.16 \\
        70 & 30    & 65.14 & 9.55 \\
        60 & 40    & 57.08 & 8.60 \\
        \bottomrule
        \end{tabular}
        \caption{Different ratios of VP and VG.}
    \end{subtable}
    \hfill
    \begin{subtable}[t]{0.33\linewidth}
    \begin{tabular}{cc|cc}
        \toprule
        \textbf{VP}(\%) & \textbf{IG}(\%) & \textbf{FVD$\downarrow$} &       \textbf{FID$\downarrow$} \\
        \midrule
        100 & 0    & 52.52 & 8.71 \\
        95 & 5    & 39.36 & 8.21 \\
        90 & 10    & \textbf{38.44} & 7.90 \\
        80 & 20    & 42.17 & \textbf{7.89} \\
        60 & 40    & 55.68 & 8.74 \\
        40 & 60    & 42.06 & 8.57 \\
        \bottomrule
        \end{tabular}
        \caption{Different ratios of VP and IG.}
    \end{subtable}
    \caption{\textbf{Ablation Studies of Multi-task.}}
    \vspace{-0.2in}
    \label{tab:mt}
\end{table*}
\begin{figure}[t]
\centering
    \includegraphics[width=1\linewidth]{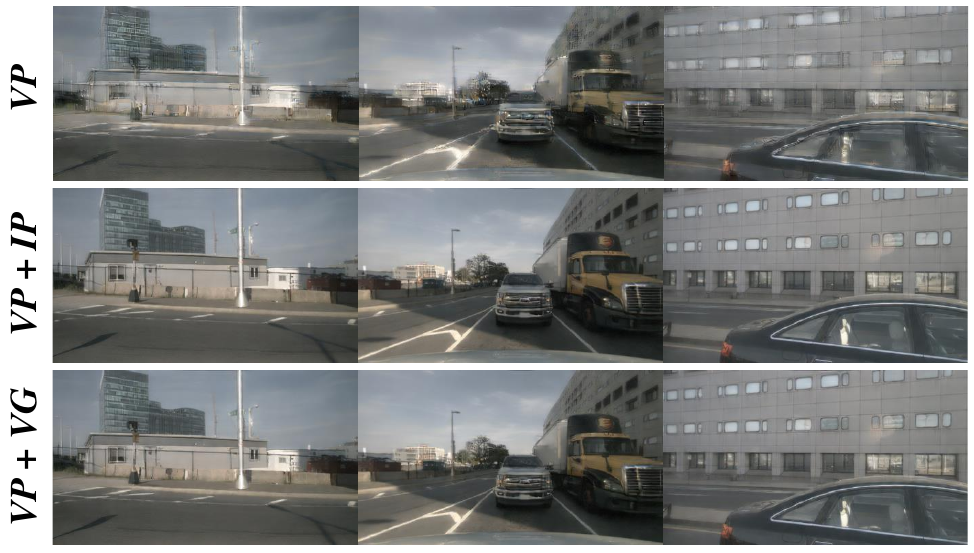}
    \caption{
    \textbf{Comparison of long-term video generation between VP, VP+IP, and VP+VG.} Introducing IP and VG tasks on top of VP can enhance the quality of frames after multiple autoregressive iterations.}
    \label{fig:abl_vp}
    \vspace{-0.2in}
\end{figure}
\noindent\textbf{Multi-task.} Tab.~\ref{tab:mt} shows the ablation study results across multiple tasks. We used VP as our primary generation task and then combined it sequentially with other tasks to determine an optimal multi-task ratio. In general, the appropriate combination of each task improves the quality of generation. Tab.~\ref{tab:mt} (a) and Tab.~\ref{tab:mt} (b) show that IP and VG primarily enhance video consistency, both applying masks on the reference frames and encourage the model to focus on using adjacent frames to ensure temporally consistent rather than relying excessively on the reference frame. Notably, introducing these two tasks greatly improves the quality of long video generation, as shown in Figure~\ref{fig:abl_vp}. Moreover, Table~\ref{tab:mt} (c) indicates that the IG task improves both video consistency and quality. By randomly dropping the temporal module, UniMLVG learn to assign different functions to different models. The temporal module focuses more on maintaining temporal consistency, while the cross-view module is dedicated to generating multi-view consistent frames.

\noindent\textbf{Multi-stage.} 
We employ a multi-stage training strategy to ensure that the model can be trained both efficiently and stably. The performance at different stages is shown in Table~\ref{tab:mts}. We find that training on large-scale single-view datasets allows the model to develop generalization and some kind of temporal generation capability. Without having seen any nuScenes data, the model achieves an FVD of $149.70$ on the nuScenes validation set, already surpassing MagicDrive~\cite{magicdrive}. This indicates that the model has already developed the generalization ability to predict future frames based on reference frames. In the second stage, the model achieves a substantial leap in both temporal consistency and image quality. In this phase, the model not only relies on temporal information and cross-view integration but also possesses condition controllability. Finally, we performed full fine-tuning on the model to unlock its potential, resulting in further improvements in the metrics.

\section{Conclusion}
\label{sec:conclusion}

To address the growing demand for generating realistic surround-view videos in autonomous driving, we propose a unified multi-view long video generation framework that supports multi-dataset training and offers versatile condition control capabilities. Specifically, we introduce a multi-task, multi-stage training strategy that effectively alleviates scene inconsistencies during long-term video generation. Moreover, by incorporating diverse datasets, image-level descriptions, and 3D conditions, our model achieves flexible control, such as generating snowy nuScenes scenes that do not exist in the original dataset. Furthermore, we are the first to introduce explicit camera viewpoint modeling for this task, which significantly enhances the consistency of video generation.
We hope that our work can contribute to advancing the development of autonomous driving.  

\clearpage
{
    \small
    \bibliographystyle{ieeenat_fullname}
    \bibliography{ref}
}

\clearpage
\setcounter{page}{1}
\maketitlesupplementary

\section{Details on Explicit Perspective Modeling}
Explicit perspective modeling aims to inject spatial information into UniMLVG to enhance the coherence of generated videos. Specifically, we utilize the camera's intrinsic parameters $K\in \mathbb{R}^{B\times T\times V \times 3 \times 3}$ and the extrinsic transformations $E \in \mathbb{R}^{B\times T\times V \times 3 \times 4}$ to obtain ray maps that match the size of the images. Importantly, we establish a unified coordinate system with the optical center of the forward-facing camera in the first frame as the origin. 
In this way, we can obtain the camera's origin coordinates $Ray\_o = E_{
4}$, where $E_{
4} \in \mathbb{R}^{B\times T\times V \times 3}$ represent the latest columns of the camera extrinsic matrices. We extend the camera origins to $\mathbb{R}^{B\times T\times V \times 3 \times H \times W}$ to match the number of pixels.
We then define a three-dimensional pixel index plane (in homogeneous coordinates) $P\in \mathbb{R}^{B\times T\times V \times 3 \times H \times W}$ with the same dimensions as the image latent space. Using the scaled camera intrinsic parameters and transformations, the ray directions from the origin to the plane can be computed as follows: $Ray\_d=E_{:3,:3}\times K^{-1}\times P$, $E_{:3,:3}$ refers to the upper-left $3\times3$ rotation matrix of the extrinsic matrix $E$. After transforming the encoded $Ray\_o$ and $Ray\_d$ through an MLP, the resulting features are added to the image latent before feeding it into the cross-view and temporal modules.

\section{Details on fusion of Cross-view and Temporal Information}
The fusion of cross-view and temporal information is essential to this task. 
We believe that the original text-to-image generation model~\cite{sd3} already possesses strong image generation capabilities, requiring only minor modifications to the image latent to achieve cross-view consistency and temporal coherence.
Therefore, we simply use a learnable parameter to perform a weighted summation of the outputs from the cross-view or temporal models with the backbone's image latent. Specifically, both the cross-view and temporal modules are GPT-2-style self-attention mechanisms~\cite{gpt2}. The fusion process can be formulated as: 
\begin{equation}
z_l = \text{Sigmoid}(\alpha) \cdot z'_l + (1-\text{Sigmoid}(\alpha)) \cdot \mathcal{F}_{l}(z'_l), 
\end{equation}
where $\alpha$ is a learnable parameter, initially set to $2$ to facilitate gradual model training, $z'_l$ denotes the output image latent from the backbone at the $l$-th layer, and $\mathcal{F}$ represents either the cross-view module or the temporal module.
In addition, we found that it is not necessary to add these two modules after every layer of the backbone; adding them at intervals does not affect performance.

\begin{figure*}[t]
\centering
    \includegraphics[width=0.9\linewidth]{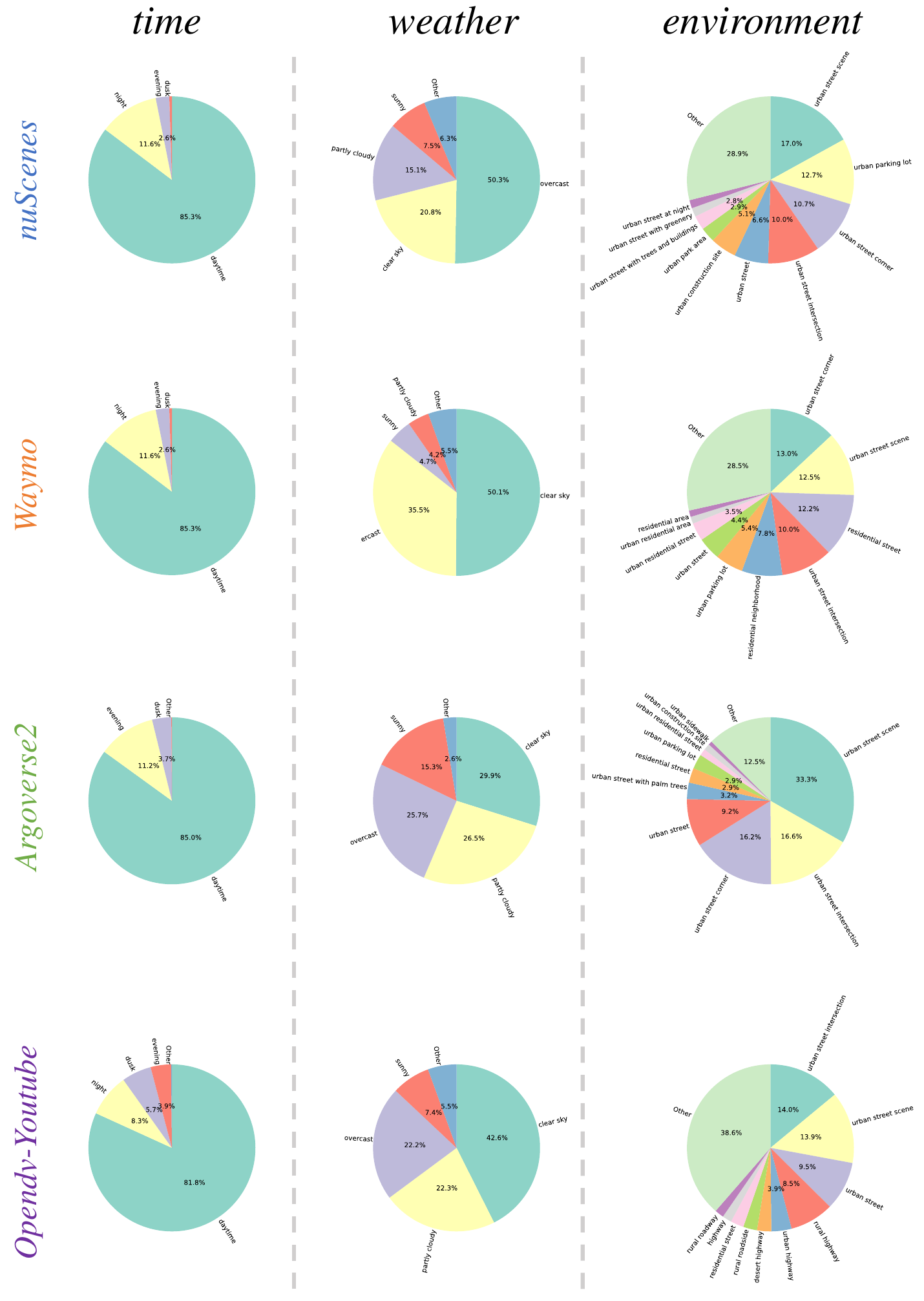}
    \caption{\textbf{Statistical Analysis of time, weather, and environment in text descriptions on four datasets}.}
    \label{fig:text_ana}
    
\end{figure*}
\section{Details on Image-level Description}
In previous works~\cite{drive-wm, magicdrive, drivedreamer2, dive}, scene descriptions were exclusively used as textual conditions for multi-view video generation. However, such descriptions often lack the fine-grained details necessary for high-quality and consistent generation. To address this limitation, we incorporate image-level descriptions, enabling more precise control over the generation process and improving the consistency and realism of multi-view videos. Specifically, we leverage the multimodal model Drivemlm~\cite{drivemlm}. For each view, we input the image along with two question prompts: "\textit{Describe the time, weather, environment, objects, and each value should be a single string and less than 20 words.}" and "\textit{Describe objects in this image within about 30 words.}" to generate detailed annotations of the time, weather, environment, and objects present in the viewpoint. Figure~\ref{fig:text_ana} presents the statistical information on time, weather, and environment annotations across the four datasets. We can observe that daytime scenes dominate across all four datasets, accounting for more than 80\% of the data. In terms of weather, the two most frequently occurring descriptors are \textit{overcast} and \textit{clear sky}. It is worth noting that snowy scenes in Argoverse2 constitute less than 2.6\%. However, UniMLVG can still modify weather text conditions to transform scenes under other weather conditions into snowy scenes, demonstrating its robust generalization and semantic understanding capabilities. In terms of environmental descriptions, we can observe the primary focus of data collection for each dataset. or instance, nuScenes, Waymo, and Argoverse2 primarily capture urban street environments, whereas OpenDV-Youtube exhibits a broader range of scenes, including highways and deserts.

\vspace{-2mm}
\section{More Qualitative Results}
\begin{figure*}[t]
\centering
    \includegraphics[width=0.95\linewidth]{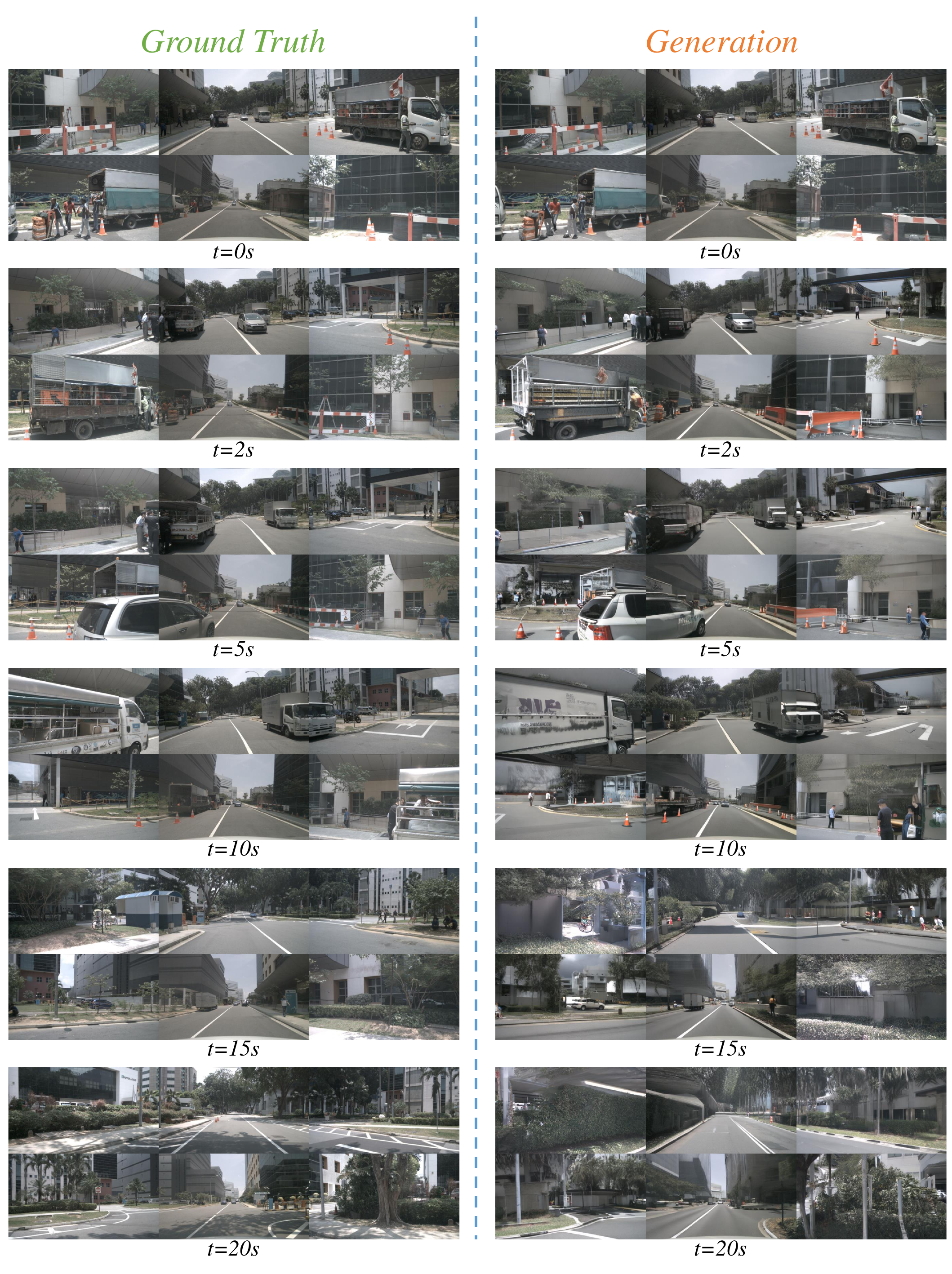}
    \caption{Sample of 20s multi-view video in a sunny scene with reference frames.}
    \label{fig:ref_sunny}
\end{figure*}

\begin{figure*}[t]
\centering
    \includegraphics[width=0.95\linewidth]{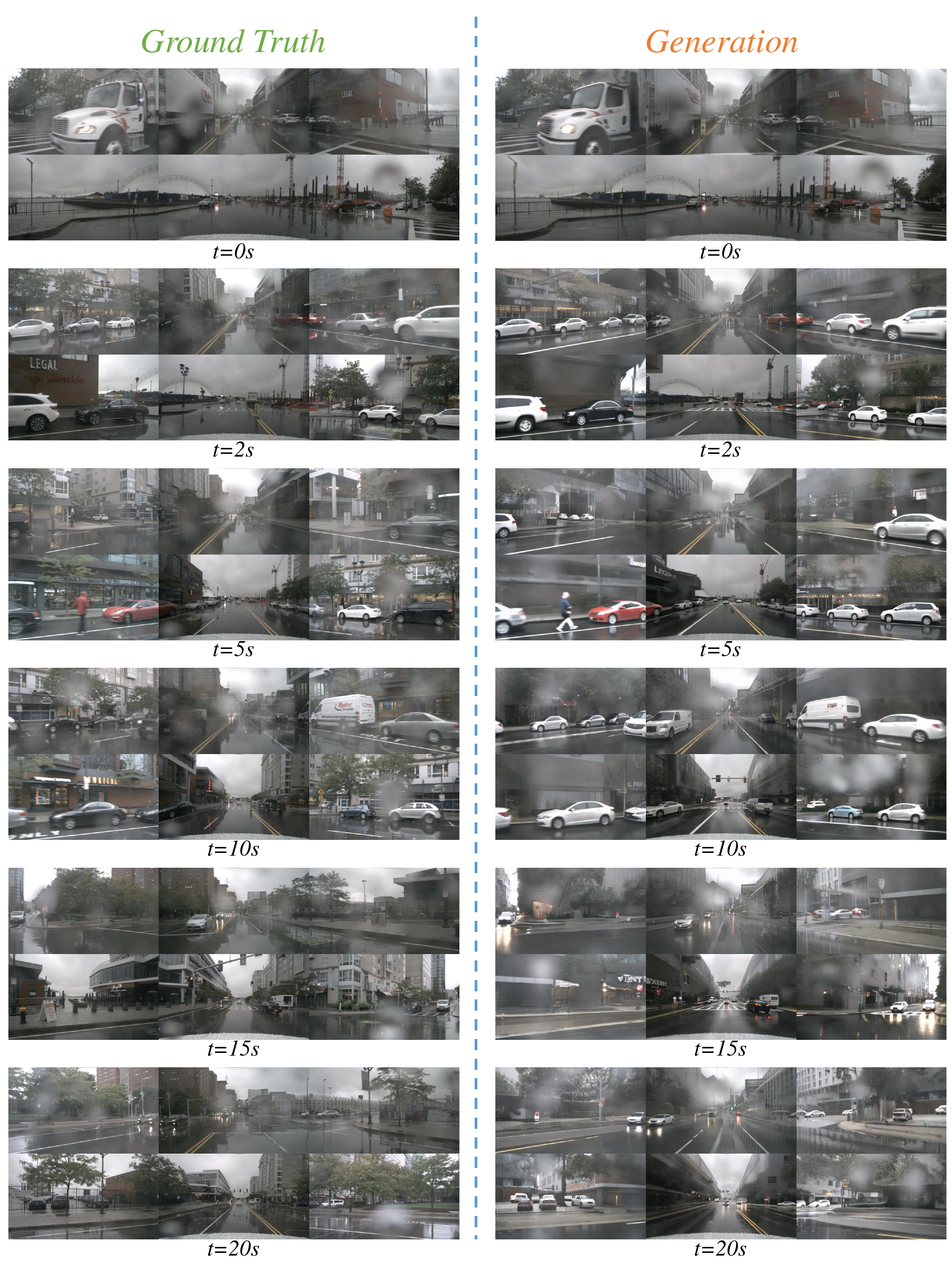}
    \caption{Sample of 20s multi-view video in a rainy scene with reference frames.}
    \label{fig:ref_rainy}
\end{figure*}

\begin{figure*}[t]
\centering
    \includegraphics[width=0.95\linewidth]{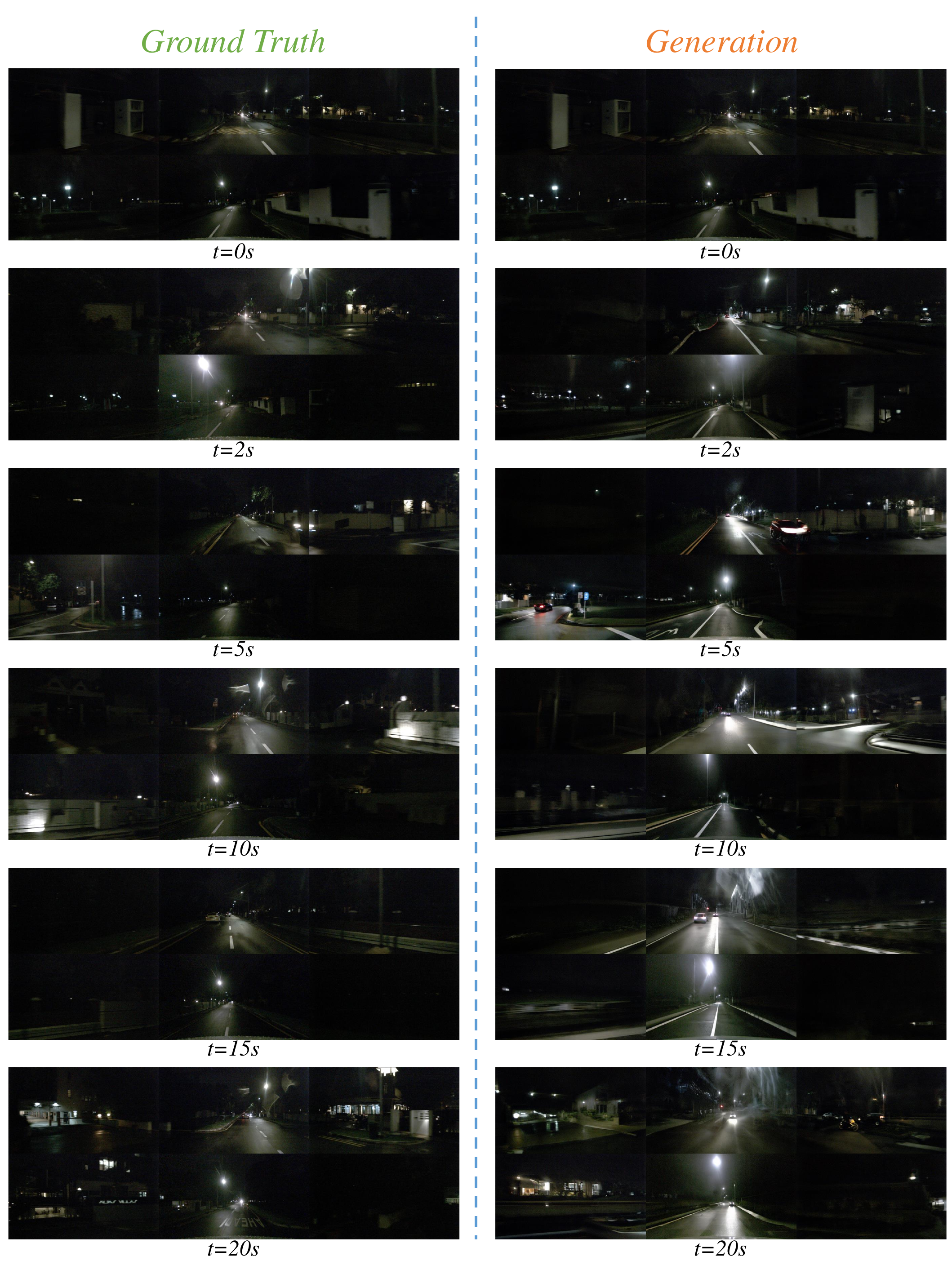}
    \caption{Sample of 20s multi-view video at night with reference frames.}
    \label{fig:ref_night}
\end{figure*}

\begin{figure*}[t]
\centering
    \includegraphics[width=0.95\linewidth]{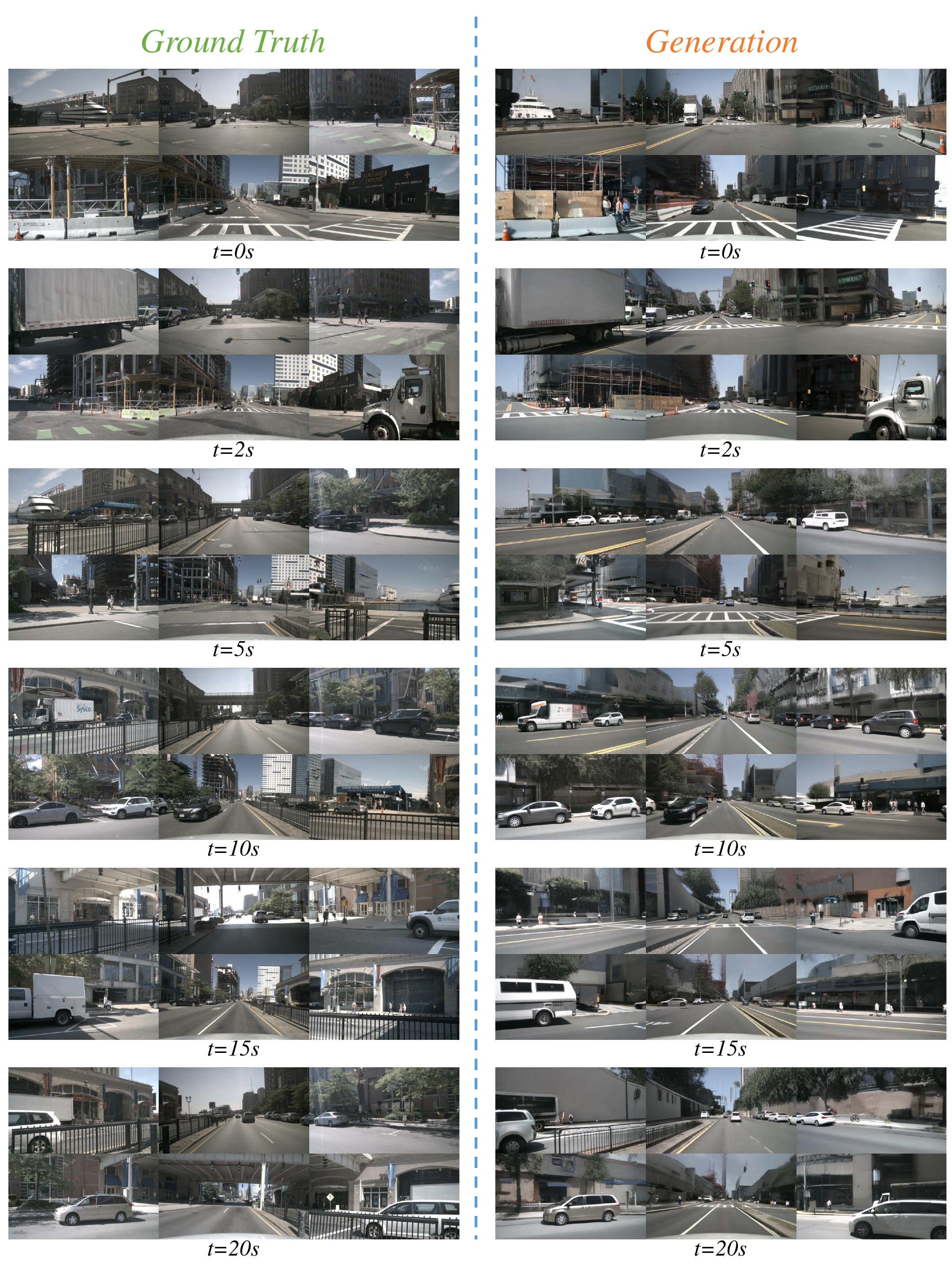}
    \caption{Sample of 20s multi-view video in a sunny scene without reference frames.}
    \label{fig:unref_sunny}
\end{figure*}

\begin{figure*}[t]
\centering
    \includegraphics[width=0.95\linewidth]{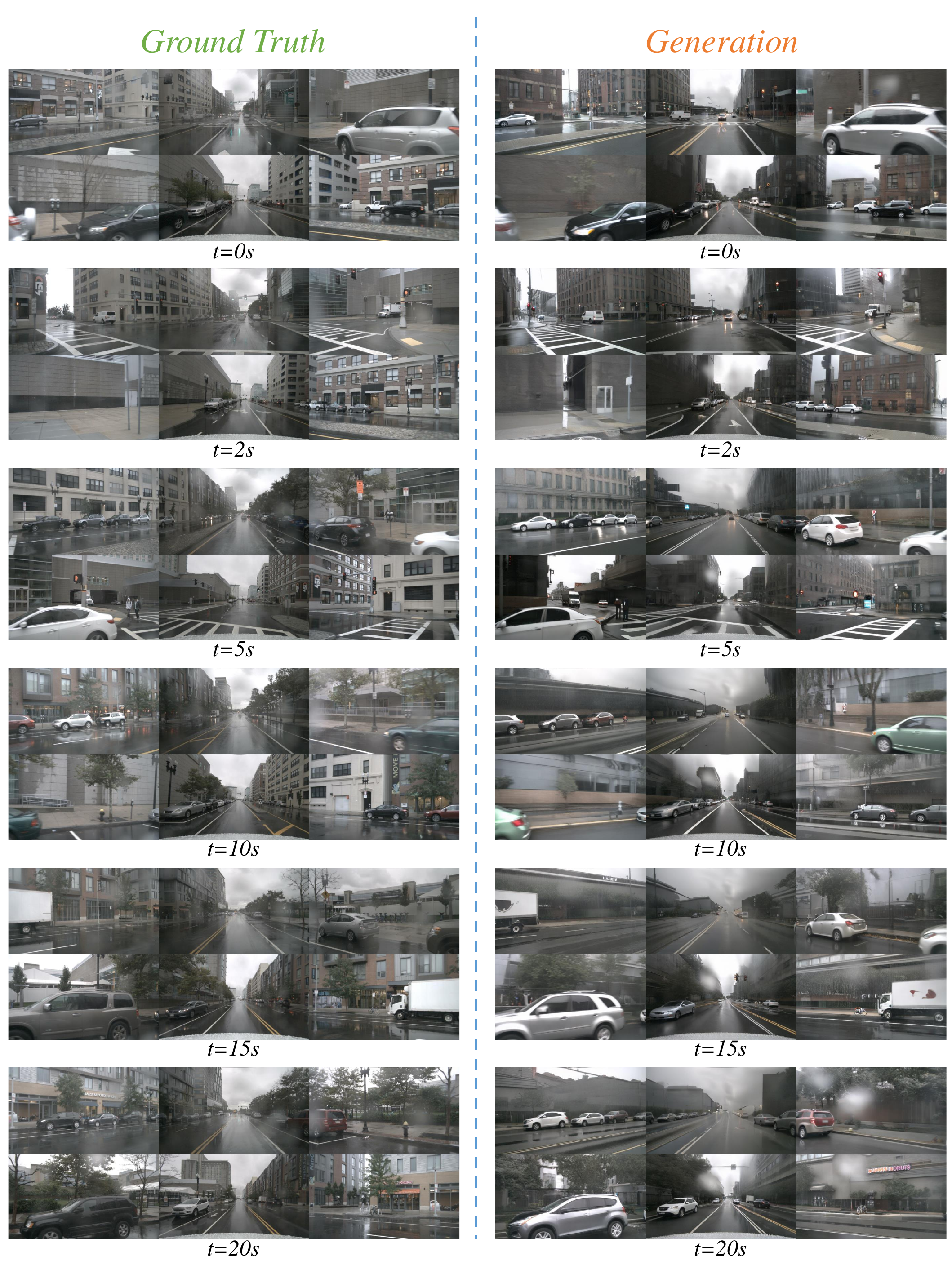}
    \caption{Sample of 20s multi-view video in a rainy scene without reference frames.}
    \label{fig:unref_rainy}
\end{figure*}

\begin{figure*}[t]
\centering
    \includegraphics[width=0.95\linewidth]{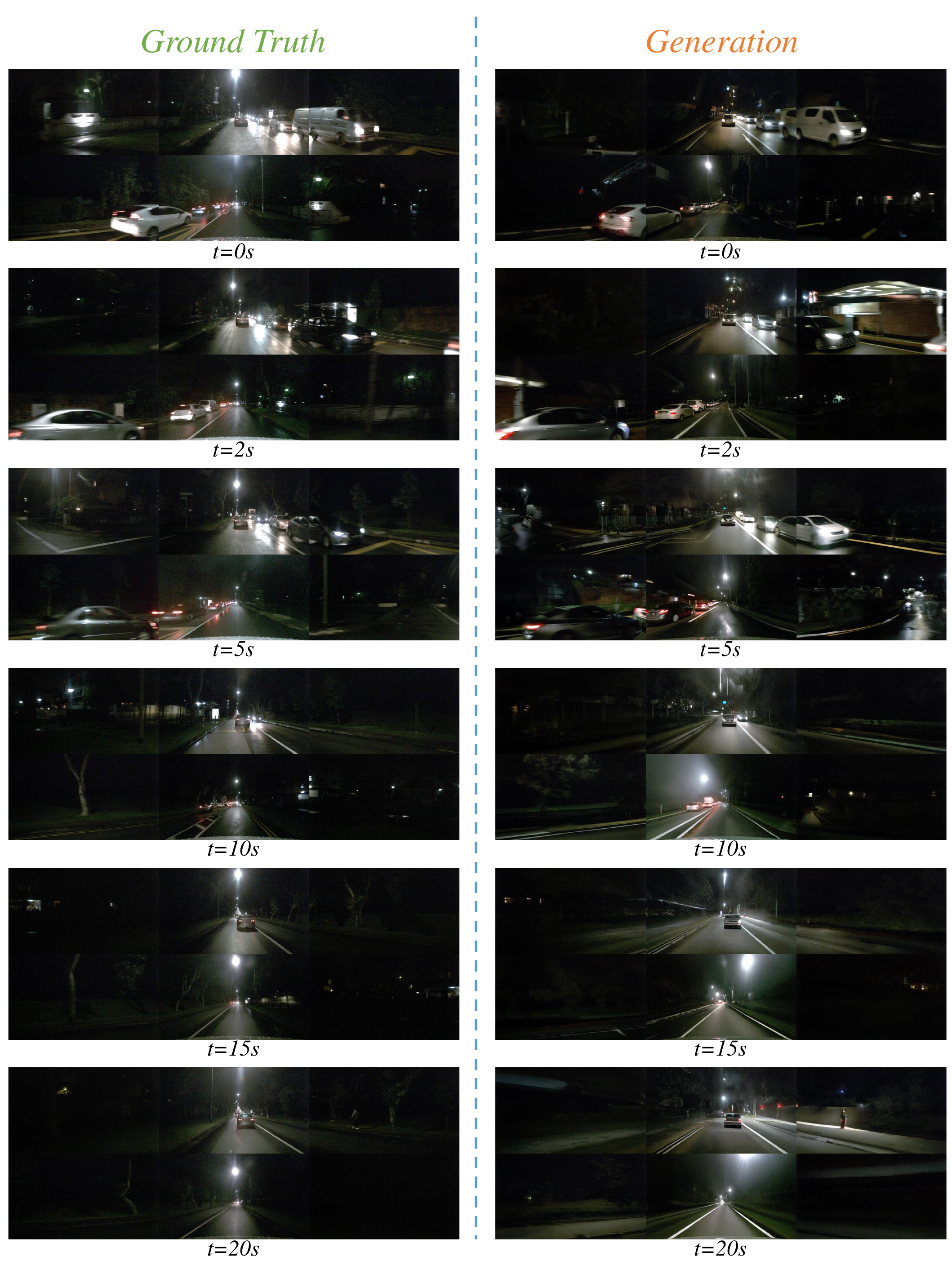}
    \caption{Sample of 20s multi-view video at night without reference frames..}
    \label{fig:unref_night}
\end{figure*}

\begin{figure*}[t]
\centering
    \includegraphics[width=0.95\linewidth]{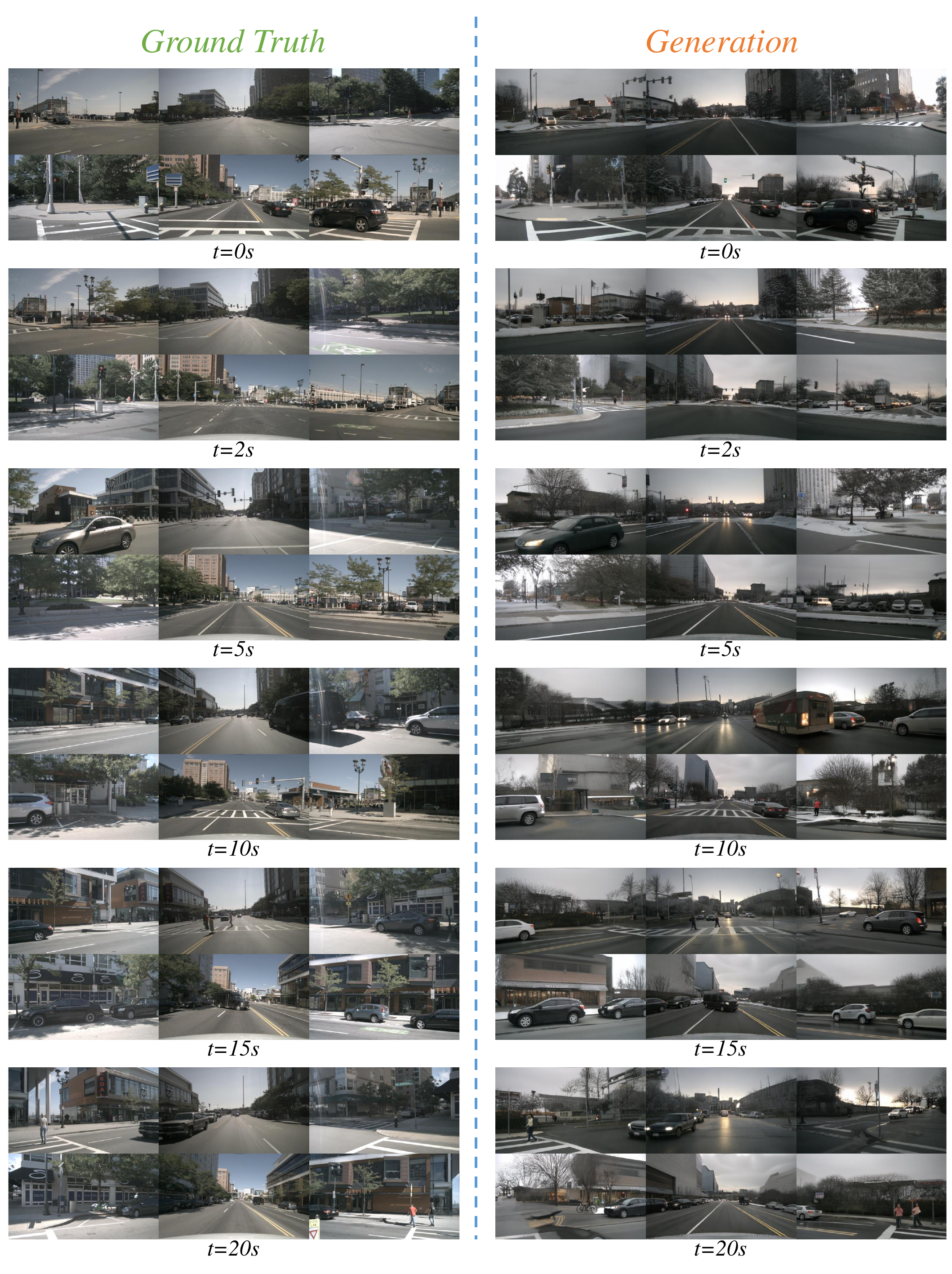}
    \caption{Sample of a 20s multi-view video transformed from a sunny to a snowy scene through text editing.}
    \label{fig:snow_sunny}
\end{figure*}

\begin{figure*}[t]
\centering
    \includegraphics[width=0.95\linewidth]{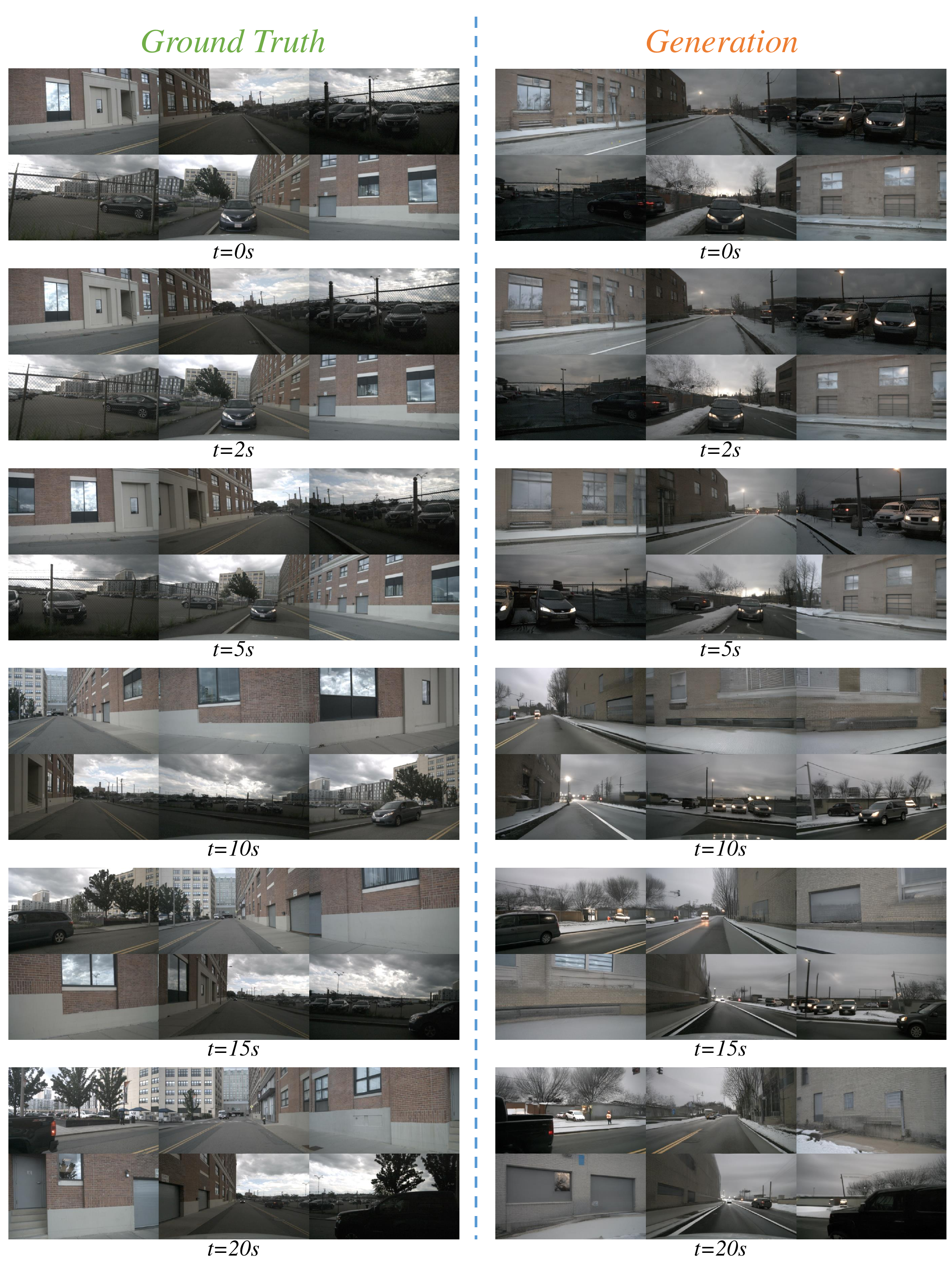}
    \caption{Sample of a 20s multi-view video transformed from a cloudy to a snowy scene through text editing.}
    \label{fig:snow_clody}
\end{figure*}

\begin{figure*}[t]
\centering
    \includegraphics[width=0.95\linewidth]{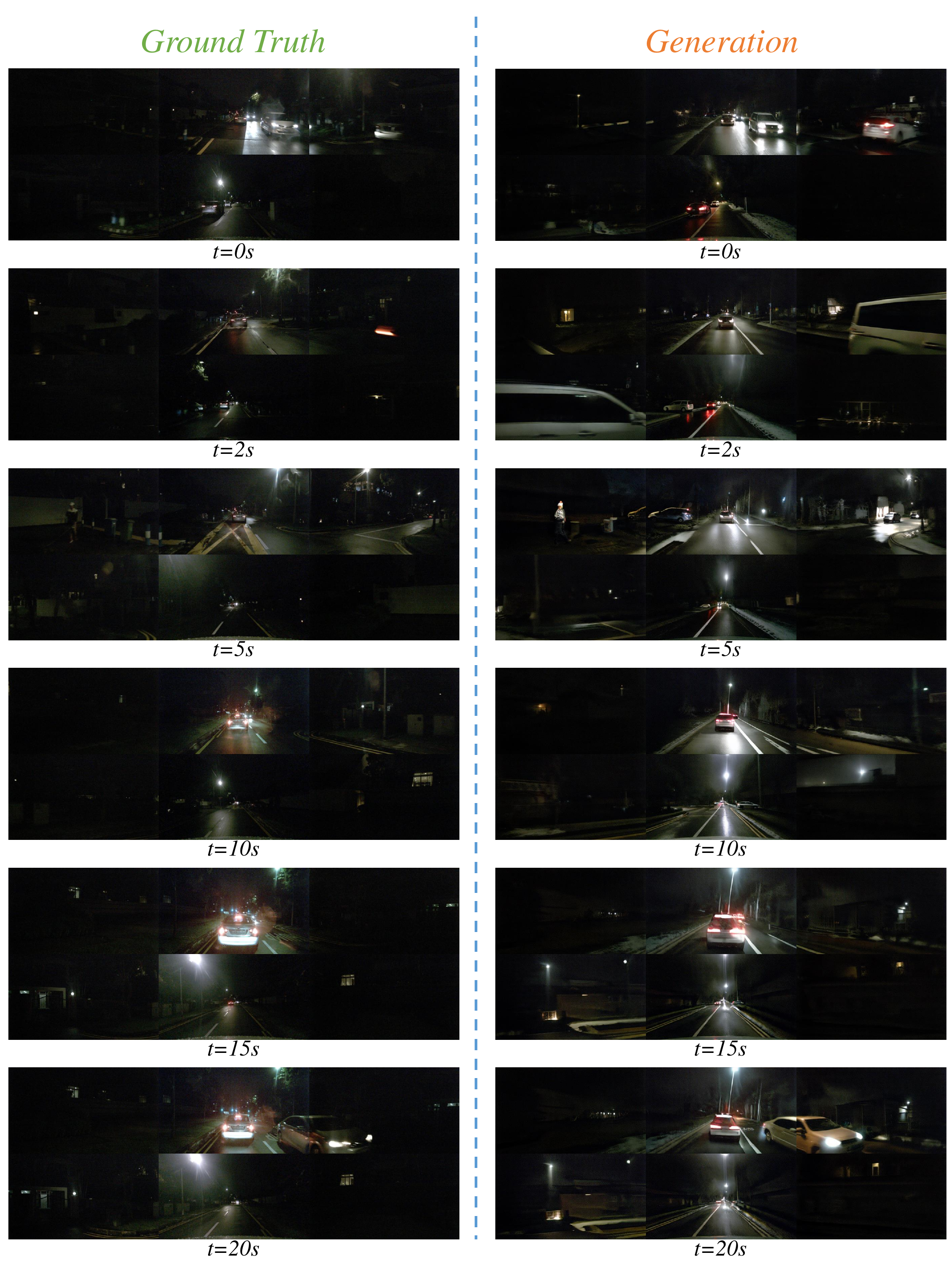}
    \caption{Sample of a 20s multi-view video generated as a snowy night scene through text editing.}
    \label{fig:snow_night}
\end{figure*}

\begin{figure*}[t]
\centering
    \includegraphics[width=0.95\linewidth]{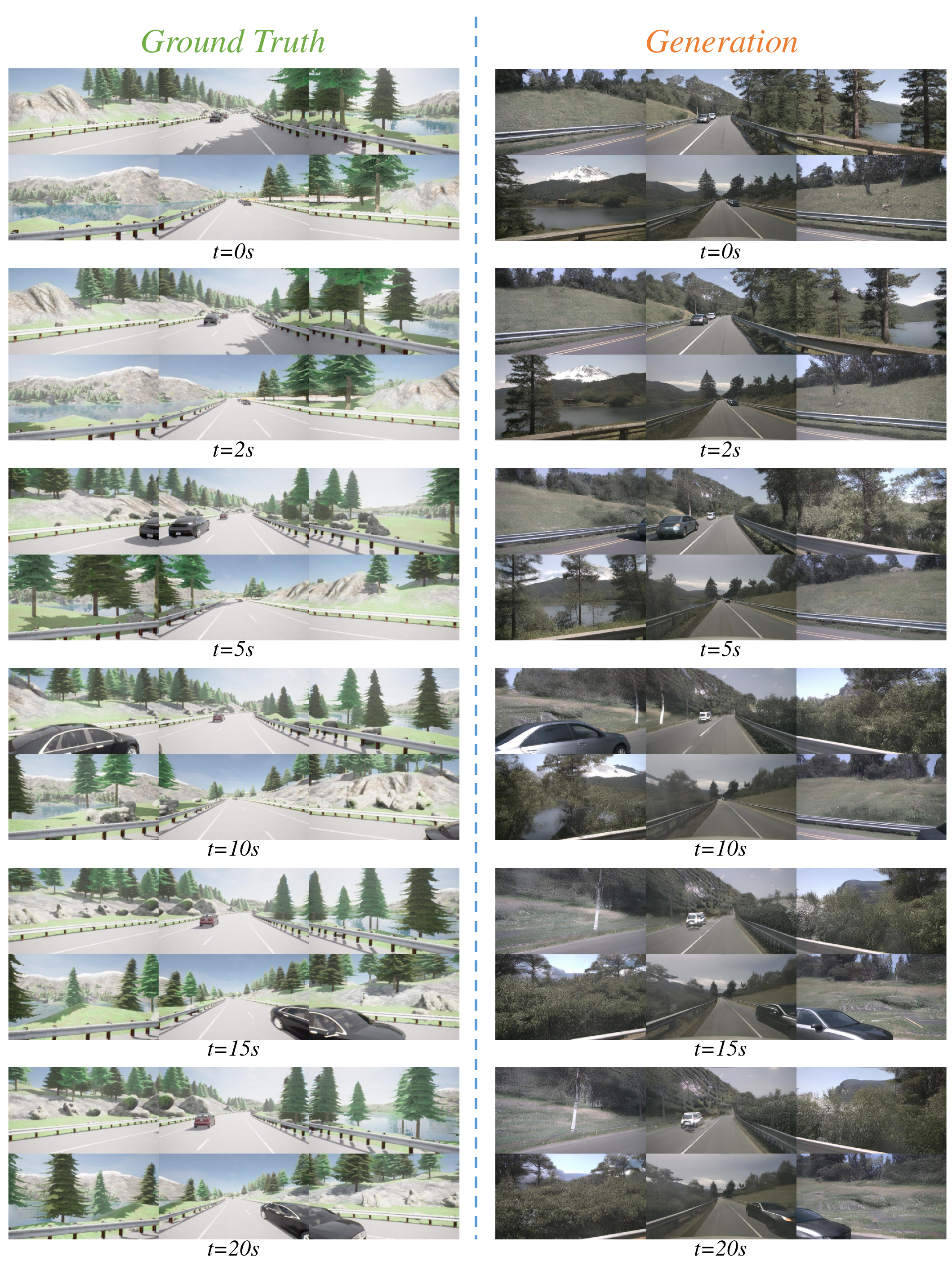}
    \caption{Sample of a realistic scene generation based on CARLA conditions.}
    \label{fig:carla04}
\end{figure*}

\begin{figure*}[t]
\centering
    \includegraphics[width=0.95\linewidth]{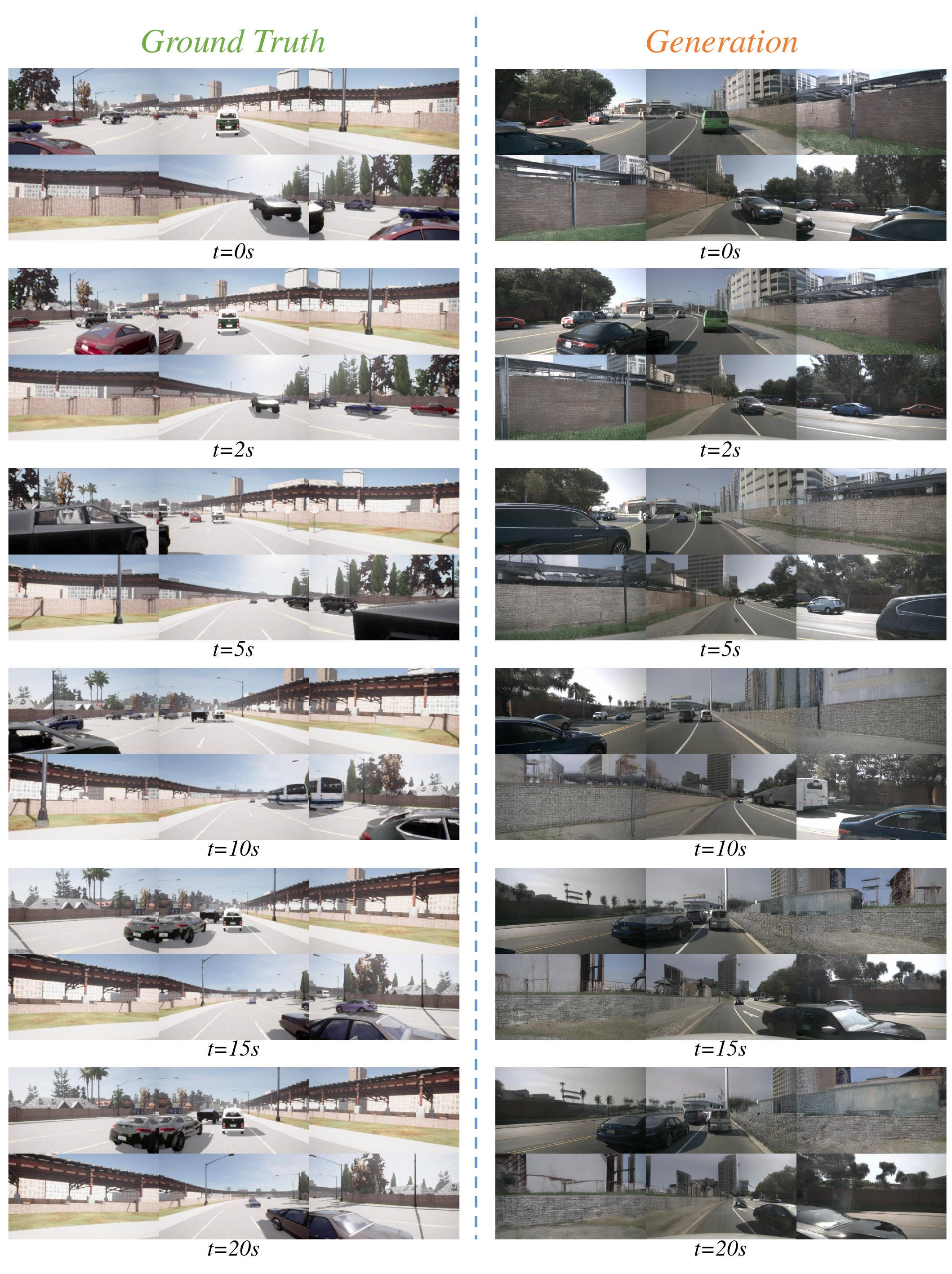}
    \caption{Sample of a realistic scene generation based on CARLA conditions.}
    \label{fig:carla05}
\end{figure*}

\begin{figure*}[t]
\centering
    \includegraphics[width=0.95\linewidth]{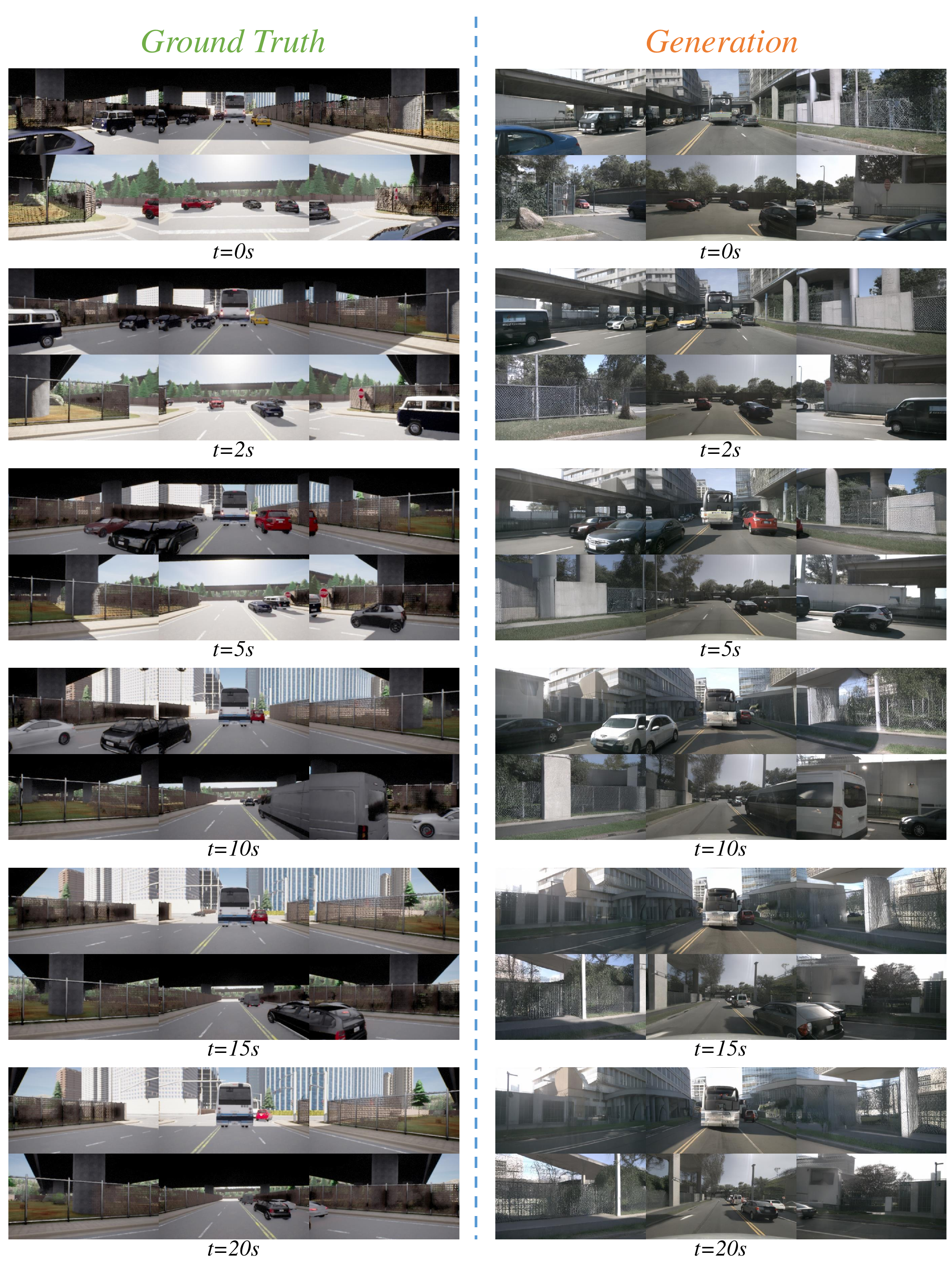}
    \caption{Sample of a realistic scene generation based on CARLA conditions.}
    \label{fig:carla10}
\end{figure*}

We provide additional examples to further demonstrate the capabilities of UniMLVG in generating long-duration, multi-view consistent videos. Figures~\ref{fig:ref_sunny}, \ref{fig:ref_rainy} and~\ref{fig:ref_night} showcase multi-view long video samples under various weather conditions and times, utilizing reference frames. Conversely, Figures~\ref{fig:unref_sunny},~\ref{fig:unref_rainy} and~\ref{fig:unref_night} illustrate multi-view long video samples under similar conditions but without reference frames. Additionally, Figures~\ref{fig:snow_sunny},~\ref{fig:snow_clody} and~\ref{fig:snow_night} demonstrate the transformation of scenes with different times and weather conditions into snowy scenes through text editing.
Finally, Figures~\ref{fig:carla04}, \ref{fig:carla05}, and \ref{fig:carla10} showcase the model's generalization capability, enabling the generation of realistic scenes based on conditions from virtual simulations.

% \clearpage
% {
%     \small
%     \bibliographystyle{ieeenat_fullname}
%     \bibliography{main}
% }
\end{document}